\documentclass[multiauthors,onecolumn]{cohere}

\title{Déjà Vu: Multilingual LLM Evaluation through the Lens of Machine Translation Evaluation}
\author{name={Julia Kreutzer},affiliation={1}}
\author{name={Eleftheria Briakou},affiliation={2}}
\author{name={Sweta Agrawal},affiliation={2}}
\author{name={Marzieh Fadaee},affiliation={1}}
\author{name={Kocmi Tom},affiliation={3}}

\affiliations{
    \item[1] Cohere Labs
    \item[2] Google
    \item[3] Cohere
}
\corresponding[*]{Julia Kreutzer (juliakreutzer@cohere.com)}
\abstract{}

\usepackage{float}
\usepackage{microtype}
\usepackage{hyperref}
\usepackage{comment}
\usepackage{url}
\usepackage{fontawesome}
\usepackage{booktabs}
\usepackage{amssymb}
\usepackage{caption}
\usepackage{multirow}
\usepackage{graphicx}
\usepackage{caption}
\usepackage{subcaption}
\usepackage{lineno}
\usepackage{tikz} 
\usepackage{xcolor}
\usepackage{enumitem}
\usepackage{marvosym}
\usepackage{colortbl}
\usepackage{titlesec}

\definecolor{rankone}{HTML}{4c7273}   % Muted teal (starting point)
\definecolor{ranktwo}{HTML}{86b9b0}   % Light teal
\definecolor{rankthree}{HTML}{a2cfc9} % Softer light teal
\definecolor{rankfour}{HTML}{c3d9d5}  % Very light teal
\definecolor{rankfive}{HTML}{d0d6d6}  % Almost pastel teal
\definecolor{ranksix}{HTML}{d6dad9}   % Lighter pastel teal
\definecolor{rankseven}{HTML}{dadbd9} % Very light pastel teal
\definecolor{rankeight}{HTML}{e0e5e5} % Faint pastel teal
\definecolor{ranknine}{HTML}{e6e9e9}  % Almost white pastel teal
\definecolor{rankten}{HTML}{f2f5f5}   % Lightest pastel teal (end point)

\definecolor{darkblue}{rgb}{0, 0, 0.5}
\hypersetup{colorlinks=true, citecolor=darkblue, linkcolor=darkblue, urlcolor=darkblue}

\definecolor{myblue}{HTML}{007BFF}
\definecolor{myred}{HTML}{DC3545}
\definecolor{mygreen}{HTML}{28A745}
\definecolor{mygray}{HTML}{6C757D}
\definecolor{mypurple}{HTML}{6F42C1}
\definecolor{myorange}{HTML}{FFA500}
\definecolor{aqua}{HTML}{00FFFF}

\newcommand{\web}{\textcolor{black}{\faGlobe}}
\newcommand{\machinegenerated}{\textcolor{black}{\faCogs}}
\newcommand{\crowd}{\textcolor{black}{\faUser}}
\newcommand{\experts}{\textcolor{black}{\faUserMd}}
\newcommand{\cmark}{\textcolor{mygreen!50}{\faCheckCircle}}
\newcommand{\xmark}{\textcolor{mygray!50}{\faTimesCircle}}
\newcommand{\lbr}{\textcolor{mygray}{[}}
\newcommand{\rbr}{\textcolor{mygray}{]}}

\definecolor{chocolate}{HTML}{D2691E} 
\newcommand{\translation}{\textcolor{chocolate}{\faLanguage}}
\newcommand{\summarization}{\textcolor{mypurple}{\faCut}}
\newcommand{\mathtask}{\textcolor{myblue}{\faCalculator}}
\newcommand{\openended}{\textcolor{mygreen}{\faMagic}}
\newcommand{\chat}{\textcolor{myred}{\faComments}}
\newcommand{\formatfollowing}{\textcolor{gray}{\faListUl}}

\newcommand{\popfirst}{\textcolor{myorange!70}{\faStar \faStar \faStar}}
\newcommand{\popsecond}{\textcolor{myorange!70}{\faStar \faStar} \textcolor{myorange!10}\faStarO}
\newcommand{\popthird}{\textcolor{myorange!70}{\faStar \faStarHalfO} \textcolor{myorange!10}\faStarO }
\newcommand{\popfourth}{\textcolor{myorange!70}{\faStar \faStar}\textcolor{myorange!10}{\faStarO}}
\newcommand{\popfifth}{\textcolor{myorange!70}{\faStar} \textcolor{myorange!10}{\faStarO \faStarO}}
\newcommand{\popsixth}{\textcolor{myorange!10}{\faStarO \faStarO \faStarO}}

\usepackage{graphicx}

\usepackage{booktabs}
\usepackage{comment}

\usepackage{listings}
\usepackage{cleveref}
\usepackage{fvextra}

\newcounter{recommendationctr}
\newcommand{\recommendation}[1]{%
  \refstepcounter{recommendationctr}%
  \smallskip\noindent\textbf{\faLightbulbO{} Recommendation~\therecommendationctr:} #1%
}

\newcommand{\mllm}{m\textsc{llm}}
\newcommand{\llm}{\textsc{llm}}
\newcommand{\mt}{\textsc{mt}}

\abstract{
Generation capabilities and language coverage of multilingual large language models (\mllm{}s) are advancing rapidly. 
However, evaluation practices for generative abilities of \mllm{}s are still lacking comprehensiveness, scientific rigor, and consistent adoption across research labs, which undermines their potential to meaningfully guide \mllm{} development.
We draw parallels with machine translation (\mt{}) evaluation, a field that faced similar challenges and has, over decades, developed transparent reporting standards and reliable evaluations for multilingual generative models. 
Through targeted experiments across key stages of the generative evaluation pipeline, we demonstrate how best practices from \mt{} evaluation can deepen the understanding of quality differences between models.
Additionally, we identify essential components for robust meta-evaluation of \mllm{}s, ensuring the evaluation methods themselves are rigorously assessed.
We distill these insights into a checklist of actionable recommendations for \mllm{} research and development.
}

\begin{document}

\section{Introduction}

Evaluating \llm{}s in a multilingual context involves testing their capabilities across different languages and tasks, with particular attention to less-studied and lower-resourced non-English languages~\citep{huang2024surveylargelanguagemodels}.
Naturally, it inherits challenges from monolingual \llm{} evaluation, such as benchmark contamination~\citep{yang2023rethinkingbenchmarkcontaminationlanguage,deng-etal-2024-investigating,dong-etal-2024-generalization,li-etal-2024-open-source,ni2025trainingbenchmarkneed}, label noise~\citep{vendrow2025largelanguagemodelbenchmarks}, costs vs coverage trade-offs
~\citep{zhang2024lmmsevalrealitycheckevaluation}, 
%coined the term ``evaluation trilemma'': ideal evaluations should be wide-coverage, low-cost, and zero-contamination, but are near possible to simultaneously to obtain.
standardization, reliability,  diversity~\citep{mcintosh2024inadequacieslargelanguagemodel}, and reproducibility issues~\citep{biderman2024lessons}. These challenges become even more evident when attempting to draw conclusions about \emph{model progress across multiple languages}.
%benchmarks lack rigor and standardization, accuracy and reliability, not representing diversity of human values and cultures and use in real-world scenarios.

Prior classification benchmarks from cross/multilingual studies that pre-date the decoder-only-\llm{} era can be re-used to gain performance insights for \mllm{}s~\citep{hu2020xtreme,ruder-etal-2021-xtreme,liang-etal-2020-xglue,ahuja-etal-2023-mega,asai-etal-2024-buffet}.
However, many of these benchmarks have reached saturation~\citep{kiela-etal-2021-dynabench,kiela2023plottingprogress} and are not separating models sufficiently~\citep{zhang2024pmmevalparallelmultilingualmultitask}. They are unreliable predictors of generative abilities of \mllm{}s~\citep{ustun-etal-2024-aya}, as they serve primarily for knowledge testing. % across languages.
Generative abilities are key in real-world applications~\citep{tamkin2024clioprivacypreservinginsightsrealworld,wu2025bitterlessonlearned2000},
%(``vibe checks'') 
and have thus moved into the spotlight of \textsc{llm}  evaluations~\citep{dubois2023alpacafarm,chiang2024chatbotarenaopenplatform,lin2024wildbenchbenchmarkingllmschallenging}. 
Multilingual models shine especially in these generative tasks, outperforming monolingual models across the bench (evidence in \Cref{app:monolingual}). However, particularly this area of evaluation is still in the early stages.

Current generative evaluation approaches for multilingual models
%(Section~\ref{sec:statusquo}) reveals that they 
\emph{lack nuances in reporting, reproducibility, standardization, robustness and reliability,} and most notably, \emph{meta-evaluation}. 
These challenges, albeit new in the \mllm{} evaluation field, are familiar problems in a sister field, the evaluation of machine translations.
In this paper, we thus establish a connection to machine translation (\textsc{mt}) evaluation research, \emph{linking new questions in \mllm{} evaluation research to known solutions in \textsc{mt} evaluation research}. 

\mt{} has had a headstart on navigating these complexities in multilingual generation evaluation.
As one of the core tasks in the \textsc{nlp} field,
it has a rich research history of evaluations
with automatic metrics~\citep{papineni-etal-2002-bleu,koehn-monz-2006-manual,lavie-agarwal-2007-meteor, 
%popovic-ney-2009-syntax,
%birch-osborne-2010-lrscore,
stanojevic-simaan-2014-beer,popovic-2015-chrf,rei-etal-2020-comet} and human judgments~\citep{vilar-etal-2007-human,birch-osborne-2010-lrscore,lopez-2012-putting,graham-etal-2013-continuous,
%sakaguchi-etal-2014-efficient, 
%graham-2015-improving, 
freitag-etal-2021-experts, kocmi-etal-2024-error},  spurred by venues like the annual Conference on Machine Translation (WMT). 
The development of evaluation went hand in hand with gradual improvement of model abilities and language coverage: Evaluation metrics that once worked sufficiently for statistical models became ineffective for neural models with superior translation quality \citep{freitag-etal-2022-results}, or for newly added languages~\citep{bapna2022buildingmachinetranslationsystems}.
Meta-evaluation~\citep[inter alia]{callison-burch-etal-2007-meta, callison-burch-etal-2008-meta, machacek-bojar-2013-results, post-2018-call, mathur-etal-2020-tangled, amrhein-etal-2022-aces, deutsch-etal-2023-ties}, i.e., the evaluation of evaluations, led to the development of evaluation and transparency standards and built a framework for metric builders.

Elements of this progress have yet to be seen in \mllm{} evaluation, due to traditionally disjoint research streams, and the rapid speed of \mllm{} development. 
To bridge this gap, we first \textbf{identify challenges} in generative \mllm{} evaluation through an assessment of current benchmarks and their adoption in model releases (\Cref{sec:statusquo}). 
 We then highlight\textbf{ five concrete evaluation principles} that are lacking in \mllm{} evaluations but established in \mt{} (\Cref{sec:adoption}). 
Finally, we establish which \textbf{prerequisites are necessary for meta-evaluations} (\Cref{sec:metaevaluation}).

We distill these findings into an \textbf{actionable checklist} for \mllm{} research (\Cref{app:checklist}),\footnote{\url{https://github.com/CohereLabs/multilingual-llm-evaluation-checklist}} to help steering \mllm{} development towards 
more reliable, expressive, and rigorous 
evaluations.

\section{The Status Quo of Multilingual LLM Generation Evaluation}\label{sec:statusquo}

\begin{table}[t]
    \centering
    \resizebox{1\textwidth}{!}{%
    \begin{tabular}{lclrlcrl|lclrlcrl}
   % \toprule
    \rowcolor{gray!10}
        \rotatebox{0}{\textbf{Benchmark}} &
        \rotatebox{0}{\textbf{Task}} &  
        \rotatebox{0}{\textbf{Rank}} &
        \rotatebox{0}{\textbf{Size}} &
        \rotatebox{90}{\textbf{Judge?}} &
        \rotatebox{90}{\textbf{Source}} &
        \rotatebox{90}{\textbf{\#Langs}} &
        \rotatebox{90}{\textbf{Transl.}} &
        \rotatebox{0}{\textbf{Benchmark}} &
        \rotatebox{0}{\textbf{Task}} &
        \rotatebox{0}{\textbf{Rank}} &
        \rotatebox{0}{\textbf{Size}} &
        \rotatebox{90}{\textbf{Judge?}} &
        \rotatebox{90}{\textbf{Source}} &
        \rotatebox{90}{\textbf{\#Langs}} &
        \rotatebox{90}{\textbf{Transl.}}\\
    \midrule

\cellcolor{chocolate!10} \textsc{flores}-$200$ & \cellcolor{chocolate!10} \translation & \popfirst & $\approx1{,}000$ & \xmark  & \web & $200$ & H & \cellcolor{myblue!10}  \textsc{mgsm} & \cellcolor{myblue!10} \mathtask & \popsecond & $250$ & \xmark  & \lbr \crowd \rbr & $10$ & H \\

  \cellcolor{chocolate!10} \textsc{ntrex}-$128$ & \cellcolor{chocolate!10}\translation & \popsixth & $\approx2{,}000$ & \xmark  &  \web & $128$ & H & \cellcolor{myblue!10}   Afri\textsc{mgsm}  & \cellcolor{myblue!10}   \mathtask & \popsixth  & $250$ & \xmark  & \lbr \crowd \rbr & $16$ & H  \\
    \cellcolor{chocolate!10}    \textsc{wmt}$24++$ & \cellcolor{chocolate!10}\translation & \popsixth  & $\approx2{,}000$ & \xmark   &  \web & $55$ & H & \cellcolor{myred!10}   SeaBench   & \cellcolor{myred!10} \chat & \popfifth & $300$ & \cmark & \experts &  $3$ & - \\
    \cellcolor{chocolate!10}    General \textsc{mt} & \cellcolor{chocolate!10}\translation & \popfourth   & $\approx2{,}000$ &  \xmark  &  \web & $\geq 11$ & H & \cellcolor{myred!10}   Sea-MTBench & \cellcolor{myred!10} \chat & \popfifth & $58$ & \cmark & \lbr \experts \rbr  & $6$ & H \\
    \cellcolor{chocolate!10}   \textsc{mafand-mt} & \cellcolor{chocolate!10}\translation & \popsixth & $1{,}000$ & \xmark  & \web & $21$ & - &  \cellcolor{mygreen!10} \textsc{mtg} &  \cellcolor{mygreen!10} \openended & \popfifth & $3{,}000$ & \cmark & \crowd/\web & $5$ & M+ \\
       % \midrule
        %\textit{Summarization} \\
     \cellcolor{purple!10}   \textsc{xls}um & \cellcolor{purple!10} \summarization & \popthird &  $500$--$11{,}000$ & \xmark & \web & $45$ & - &  
 \cellcolor{mygreen!10} \textsc{omge}val &  \cellcolor{mygreen!10} \openended & \popsixth &  $804$ & \cmark & \lbr\machinegenerated\rbr & $5$ & M \\
    \cellcolor{purple!10}     CrossSum-In  & \cellcolor{purple!10} \summarization & \popsixth & $500$ & \xmark & \lbr\web\rbr & $29$ & H &  \cellcolor{mygreen!10} mArenaHard &  \cellcolor{mygreen!10} \openended & \popfifth & $500$ & \cmark & \lbr\crowd\rbr & $23$ & M\\  
    \cellcolor{gray!10}    SEA-IFEval & \cellcolor{gray!10}  \formatfollowing & \popfifth & $105$ & \xmark &  \lbr\machinegenerated\rbr & $6$ & H &   \cellcolor{mygreen!10} Dolly translated &  \cellcolor{mygreen!10} \openended & \popfourth & $200$ & \cmark & \lbr\crowd\rbr & $101$ & M+\\ 
      \cellcolor{gray!10}   MIFEval & \cellcolor{gray!10}  \formatfollowing & \popsixth & $96$ &  \xmark & \lbr\machinegenerated\rbr & $10$ & M &  \cellcolor{mygreen!10} Aya human-ann. &  \cellcolor{mygreen!10}\openended & \popfifth & $250$ &  \cmark & \crowd & $7$ & - \\
     \cellcolor{gray!10}   MultiIF & \cellcolor{gray!10}  \formatfollowing & \popsixth & $454$--$909$ & \xmark  &  \lbr\machinegenerated\rbr & $7$ & M & \cellcolor{mygreen!10} PolyWrite & \cellcolor{mygreen!10} \openended & \popfifth &  $\approx155$  & \cmark & \machinegenerated & $240$ & M \\
       & & & & & &  & & \cellcolor{mygreen!10} MultiQ &  \cellcolor{mygreen!10}\openended & \popsixth & $200$ & \cmark & \machinegenerated/\crowd & $137$ & M\\ % https://huggingface.co/datasets/caro-holt/MultiQ
    %\bottomrule
    \end{tabular}%
    }
    \caption{Public generative benchmarks for downstream text-based evaluation of \mllm{}s. They are sourced from the web (\web), crowds (\crowd), experts (\experts), or machine generated (\machinegenerated). 
    %, with test set sizes, languages, source and translation information.  
    Brackets indicate extension of previous benchmarks. 
    %When a benchmark extends a previous one, we indicate this in brackets. 
    Translations of prompts are denoted as H(uman) and M(achine), with M+ indicating human post-edits. We mark \textsc{llm} judged benchmarks (\cmark), and rank them by popularity \popfifth in model releases, based on a survey of benchmark adoption detailed in \cref{app:model_release_benchmarks}. \Cref{tab:eval_benchmarks_full} provides more details for each of the listed benchmarks.
    %TODO: add the last column back; consider sorting by sizw or langs within each group
    }
    \label{tab:eval_benchmarks}
\end{table}

We compile a non-exhaustive list of open multilingual generative benchmarks in Table~\ref{tab:eval_benchmarks} to survey the \mllm{} landscape.\footnote{We exclude classification benchmarks such as MCQA problems, see discussion in ~\Cref{app:in_disguise}.}  We summarize trends as follows, and link (\ForwardToIndex) them to proposed strategies from \mt{} evaluation research in \Cref{sec:adoption} and meta-evaluation in \Cref{sec:metaevaluation}.

\smallskip\noindent
\textbf{Multilinguality via translation} Most tasks rely on the translation of the original English benchmark for multilingual expansion. Only \emph{XLSum}~\citep{hasan-etal-2021-xl}, \emph{Aya human-annotated}~\citep{singh-etal-2024-aya}, \emph{SeaBench}~\citep{zhang2024seallms3openfoundation}, \emph{MAFAND-MT}~\citep{adelani-etal-2022-thousand} are 
directly curated in the target languages. 
Automatic prompt translations might not be universally applicable or high-quality~\citep{zhang-etal-2023-dont,plaza2024spanishllmbenchmarksmmlu,agrawal-etal-2024-translation,thellmann2024multilingualllmevaluationeuropean}, which some benchmarks address with post-editing or localization (\textit{e.g.} \emph{SEA-IFEval}~\citep{tjhi-etal-2023-sea}). 
While translation achieves a broad coverage of languages, it limits the cultural representativeness and might propagate Western-centric and Anglo-centric biases~\citep{singh2024globalmmluunderstandingaddressing,guo2024largelanguagemodelsenglish}
(\ForwardToIndex ~\Cref{sec:translationese}).

\smallskip\noindent
\textbf{Small and not so mighty} The majority of test sets contain less than $500$ prompts per language, with \mt{} benchmarks as outliers with over $1{,}000$ samples. While prompt sourcing is a challenging task, especially with experts, such small sets raise questions of statistical power (
\ForwardToIndex ~\Cref{sec:statistical_power}). 
When included, human evaluations tend to cover even fewer instances~\citep{gehrmann2023repairing}.
%, which will be illustrated in Section~\ref{sec:statistical_power}. 
Most benchmarks provide only a test split, lacking development sets for tuning, which increases the risk of overfitting and diminishes the significance of reported improvements over time~\citep{van-der-goot-2021-need,Ott_2022}.
Qualitative insights beyond aggregated task metrics are rarely included in evaluation reports (\ForwardToIndex ~\Cref{sec:analyses}).

\smallskip\noindent
\textbf{Divergences in benchmark adoption and reporting}
Only few generative benchmarks are well-established, i.e.,  multiple labs use them for reporting results in open \mllm{} releases (\Cref{app:model_release_benchmarks}). 
Flores-200~\citep{nllb2022}, MGSM~\citep{shi2023language-mgsm}, and XLSum~\citep{hasan-etal-2021-xl} are the most popularly used benchmarks, as indicated by the rank in \cref{tab:eval_benchmarks}. 
These are closed generative evaluation tasks that have the advantage of having relatively well-defined evaluation paradigms. Open generation tasks like chat and open-ended QA have less standardized evaluations and tend to rely on \llm{} judges, which introduces more ambiguities.
What complicates cross-paper comparisons even when using the same benchmark, is the lack of transparency and standardization in evaluation reporting.
This goes from the choice of automatic metric (or \llm{} judge), over prompting conditions and formulations (\ForwardToIndex ~\Cref{sec:reproducibility}), to the selection and aggregation across languages for comparison (\ForwardToIndex ~\Cref{sec:aggregation}). 
For instance, performance on Flores-200 is measured with different metrics (spBLEU~\citep{goyal-etal-2022-flores}, ChrF~\citep{popovic-2015-chrf}, COMET-22~\citep{rei-etal-2022-comet}), and for MGSM model reports vary the number of shots, or even define new criteria \citep{salamandra}. When metrics are established for a generative task in English, they might not transfer equally well to all evaluated languages~\citep{gehrmann2023repairing}.
Sometimes, it is not even stated which languages of a benchmark are chosen for evaluation, and rarely do they cover all of the supported languages of a model (see \cref{tab:eval_benchmark_use}).

\smallskip\noindent
\textbf{Generative models are becoming the metric} 
The emergence of new generative tasks, such as chat and open-ended generation, do not come with decades of task-specific metrics research. Thus, \textsc{llm} judges are used to express preferences through pairwise comparisons of model outputs, both in training~\citep{lee2024rlaif}, and in evaluation~\citep{NEURIPS2023_91f18a12,gu2025surveyllmasajudge}.
However, this in itself is a generative evaluation task measuring how good \mllm{}s are at judging multilingual generations~\citep{gureja2024mrewardbenchevaluatingrewardmodels, 
doddapaneni2024crosslingualautoevaluationassessing} \---\ raising questions about biases in used judges \citep{ye2024justice} and gameability~\citep{zheng2025cheating} (
\ForwardToIndex ~\Cref{sec:need_for_metrics}).

\smallskip\noindent
\textbf{Evaluation with a rapidly moving target} 
All of today's leading \mllm{}s and most of the generative benchmarks are less than a year old. Benchmarks quickly ``expire'', due to score saturation as a consequence of overfitting, evolved capacities or contamination~\citep{ahuja2024contaminationreportmultilingualbenchmarks}, or they simply lose relevance for \textsc{llm} user or broader research needs~\citep{zheng2023judging, tamkin2024clioprivacypreservinginsightsrealworld,wu2025bitterlessonlearned2000}. 
Many new model releases include testing on newly introduced benchmarks to highlight new strengths, but these benchmarks are rarely adopted by consequent releases of other labs.
Open leaderboards attempt to close this gap, tracking progress on a selection of tasks and languages across models (see \Cref{sec:leaderboards}), but they are also prone to expiry, might lack utility~\citep{ethayarajh-jurafsky-2020-utility} and heavily rely on aggregations for interpretation, which requires particular care for multilingual models~\citep{cohereblog} (\ForwardToIndex ~\Cref{sec:aggregation}). 
This calls for a larger arc of evaluation, namely the evaluation of evaluations themselves, including automatic (\ForwardToIndex ~\Cref{sec:need_for_metrics}) and human evaluation
(\ForwardToIndex ~\Cref{sec:need_for_human}).

\section{Adopting Evaluation Practices from MT Evaluation}\label{sec:adoption}

Based on the challenges outlined in \Cref{sec:statusquo}, we identify five central questions in the  \mllm{}
evaluation pipeline and relate them to insights and practices from \mt{}. 
The guiding question is: 
\emph{What knowledge would we gain about \mllm{}s, if we supplemented their evaluations with \mt{}-style evaluation techniques?}

\subsection{Where Does the Data Come From? Treating Synthetic Data with Care}\label{sec:translationese}

Machine translated datasets are commonly used in \mllm{} training~\citep{dang-etal-2024-rlhf} and evaluation~\citep{lai-etal-2023-okapi}, 
with the intention to reduce data scarcity across languages~\citep{muennighoff-etal-2023-crosslingual,holmstrom-doostmohammadi-2023-making,ustun-etal-2024-aya}. 
However, synthetic, model-generated data is prone to systematic biases~\citep{ahn-etal-2022-knowledge,lukasik2022teachers,shimabucoro-etal-2024-llm}.
In particular, machine-translated prompts may contain translation artifacts affecting evaluation outcomes~\citep{chen-etal-2024-good-data,guo2024largelanguagemodelsenglish,agrawal-etal-2024-translation}. In \textsc{mt} research, studies have shown that grammar, structure, or word choice of the source text can systematically influence human and machine translations
%, often resulting in less natural output 
\---\ a phenomenon known as \emph{translationese}~\citep{gellerstam1986translationese,laviosa2011corpus}. In evaluations, the presence of translationese in the sources 
(i.e., because the source text was itself translated from the target language) 
has been found to decrease the difficulty of the task~\citep{zhang-toral-2019-effect}, even leading to false claims of human parity~\citep{hassan2018achievinghumanparityautomatic,toral-etal-2018-attaining,graham-etal-2020-statistical}.
%\---\ which were later debunked upon correction of the test sets' directionality~\citep{toral-etal-2018-attaining,graham-etal-2020-statistical}. 
As a result, maintaining source authenticity has become a critical principle in the creation of test sets~\citep{barrault-etal-2019-findings}.

 \faFlask{}  To illustrate the effects of prompt translation in multilingual generative evaluation, we conduct an experiment using $250$  {\href{https://huggingface.co/datasets/CohereLabs/aya_evaluation_suite/viewer/aya_human_annotated/test?sort[column]=language&sort[direction]=asc}{Aya human annotated} prompts for Arabic, Chinese, English, Portuguese and Turkish.
 We round-trip-translate them automatically via a pivot language to create a comparison between original prompts and translated prompts~\citep{chen-etal-2024-good-data}. For translation, we use a diverse range of translators: Google Translate (\textsc{gt}), and \textsc{nllb-200-3.3B}~\citep{nllbteam2022languageleftbehindscaling}, \textsc{Aya Expanse 32B}~\citep{dang2024ayaexpansecombiningresearch} and \textsc{Command A}~\citep{cohere2025command}.
 We measure how GPT-4o-as-a-judge win rates change when comparing \mllm{} generations for original prompts to those for translated prompts. 
 We focus on a comparison of %\textsc{LLama3.1 8B} Instruct~\citep{grattafiori2024llama3herdmodels}, 
 \textsc{Gemma2 9B}~\citep{gemmateam2024gemma2improvingopen} and \textsc{Aya Expanse 8B} ~\citep{dang2024ayaexpansecombiningresearch}, 
selected from a wider range of experiments in ~\Cref{app:translation}.

\begin{figure}[t!]
\begin{minipage}[t!]{0.55\textwidth}
     \includegraphics[width=\linewidth]{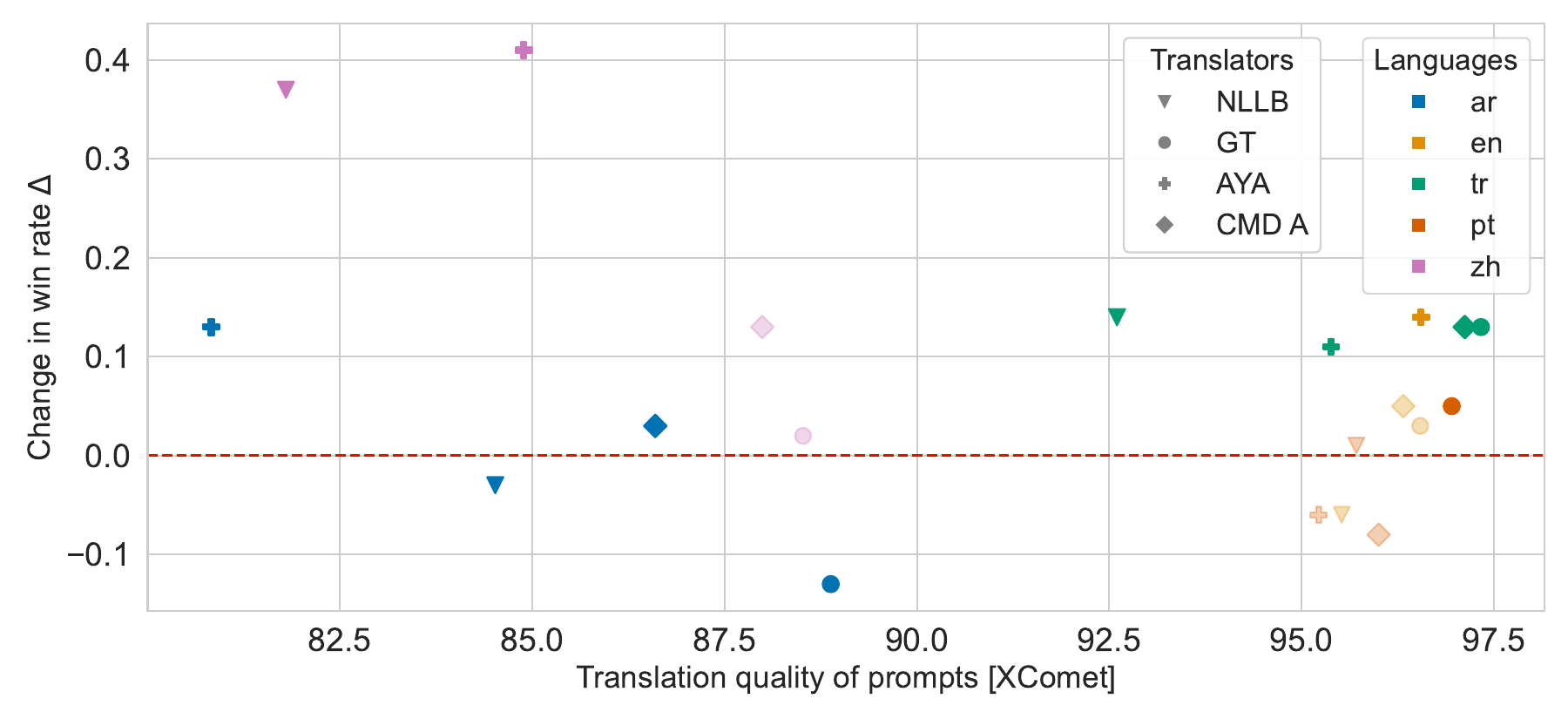}
    \captionof{figure}{The effect of prompt translation quality on win rates differences between \textsc{Aya Expanse 8B} vs \textsc{Gemma2 9B}: Win rate $\Delta$s mostly increase compared to the ones under original prompts ($y=0$). Transparent points reflect non-significant win rate $\Delta$s (at 95\% CI). 
    }
    \label{fig:translation_effects}
\end{minipage}
\hfill
\begin{minipage}[t!]{0.4\textwidth}%\vspace{0pt}%
%\begin{table}[t]
\centering
\resizebox{0.9\textwidth}{!}{
\begin{tabular}{lcc}
\toprule
\textit{Average:} & \textsc{XComet} $\uparrow $& Change in Win Rate $\Delta$ $\downarrow$\\
 \midrule
     \textsc{nllb} & 90.03& 0.06 \\
     \textsc{gt} & 93.65 & 0.02 \\
     \textsc{aya} & 90.57 & 0.14 \\
     \textsc{Cmd A} & 92.80 & 0.05 \\
     \bottomrule
\end{tabular}%
}
\captionof{table}{Average roundtrip translation quality of translation models tested on \textit{Aya human annotated} prompts across five languages (ar, en, pt, tr, zh), and the change in win rate $\Delta$ when comparing \textsc{Aya Expanse 8B} and \textsc{Gemma 2 9B}. Ideally, translation should not affect the win rate. }
\label{tab:avg_mt_quality}
%\end{table}
\end{minipage}
\end{figure}

\begin{comment}
    
\begin{figure}
    \centering
    \includegraphics[width=0.6\linewidth]{figures/translation_quality_vs_wr_scale.png}
    \caption{The effect of prompt translation quality on change in win rates when comparing \textsc{Aya Expanse 8B} to \textsc{Gemma2 9B}. The red line indicates the win rates of original prompts. }
    \label{fig:enter-label}
\end{figure}
\end{comment}

\faSearch{} 
We find that \textbf{win-rate differences in pairwise evaluations are affected by translation, with magnitudes that vary across languages and translation models} (\cref{fig:translation_effects}), depending on translation quality (\cref{tab:avg_mt_quality}).  
We can see that the majority of translations tilt the scale in favor of \textsc{Aya Expanse 8B}, increasing the win rate delta over \textsc{Gemma2 9B}
%Translations by the larger \textsc{Aya Expanse 32B} model give \textsc{Aya Expanse 8B}'s win-rates a substantial 
from $0.18$ to $0.32$ on average across languages (especially for Chinese and Turkish).
Why hypothesize that \textsc{Aya Expanse 8B} is more robust to translation artifacts in the prompts due to exposure during training~\citep{artetxe-etal-2020-translation}.
We also note that \mllm{}s used for translation (\textsc{Aya Expanse 32B}, \textsc{Command A}) appear to have proportionally larger downstream effects.
Overall, this simulation demonstrates that win rates computed on translated evaluation prompts might systematically favor particular models that are more robust to translation artifacts, leading to inflated win rates.

\recommendation{For evaluation, prefer target-language original prompts over translated alternatives (\emph{silver standards}, coined by~\citet{holtermann-etal-2024-evaluating}).
% , as translations can influence model rankings. 
If translations are unavoidable, ensure that their quality is optimized without assuming off-the-shelf adequacy for any task (\textit{e.g.} choosing best \mt{}, adding post-edits, localization).
Measure and document translation quality on a representative subset for each task.}

\subsection{What do Score Differences Mean? Measuring Significance, Power \& Effect Size}\label{sec:statistical_power}

Although platforms like Chatbot Arena report confidence intervals using bootstrapping, significance testing is not yet a standard part of the \textsc{llm} development pipeline~\citep{vaugrante2024loomingreplicationcrisisevaluating, ackerman2025statisticalmultimetricevaluationvisualization}. To address this, \citet{miller2024addingerrorbarsevals} proposed best \textsc{llm} evaluation practices, emphasizing the importance of reporting sample size, confidence intervals, and standard errors, particularly for clustered and paired tests. In \mt{} research, such reporting and significance tests have a long history~\citep{koehn-2004-statistical, riezler-maxwell-2005-pitfalls, graham-etal-2014-randomized, graham-etal-2020-statistical} and have found moderate adoption~\citep{marie-etal-2021-scientific}, also enabled by ease of use in tools like \texttt{sacrebleu}~\citep{post-2018-call} or \texttt{comet-compare}~\citep{rei-etal-2020-comet}.
%\footnote{\url{https://unbabel.github.io/COMET/html/index.html}} 
Statistical power analyses can further help determine the sample size required for reliable evaluations~\citep{card-etal-2020-little}, \textit{e.g.} for human preference evaluation. In \mt{}, for instance, statistical power is usually sufficient ($>0.8$) to rank even close models with $\approx$1.5K sources~\citep{graham-etal-2020-statistical}, yet smaller sample sizes may be insufficient \citep{wei-etal-2022-searching}.
Based on test sizes of the benchmarks reviewed in \Cref{sec:statusquo}, it is likely that especially under metrics with high variance such as pairwise LLM judgments, many \mllm{} evaluations might be underpowered.

There are some pitfalls to be aware of: first, different metrics for the same task may have varying sensitivity~\citep{riezler-maxwell-2005-pitfalls}, which could lead to differences in one metric being significant but insignificant in another. Second, the more statistical tests are done, the more likely false positives will be encountered. This becomes particularly relevant for testing multiple models on multiple languages and multiple benchmarks. Correction~\citep{zerva-etal-2022-findings,ulmer2022deep} can prevent this inflation, \textit{e.g.} by increasing the threshold of significance for individual tests (Bonferroni correction), implemented at WMT.

It is important to recognize that statistical significance does not necessarily imply that a difference is noticeable or meaningful to humans~\citep{mathur-etal-2020-tangled,agrawal-etal-2024-automatic-metrics}. With sufficiently large sample sizes, even very small differences in metric scores can become statistically significant, despite being too subtle to notice in practice. This issue is specifically known in \mt{}, where the magnitude of the effect size plays a crucial role in determining whether system improvements are genuinely meaningful~\citep{kocmi-etal-2024-navigating}.

\begin{figure}[t!]
  \centering
\begin{minipage}[t]{0.55\textwidth}%\vspace{0pt}%
\centering
 \includegraphics[scale=0.3]{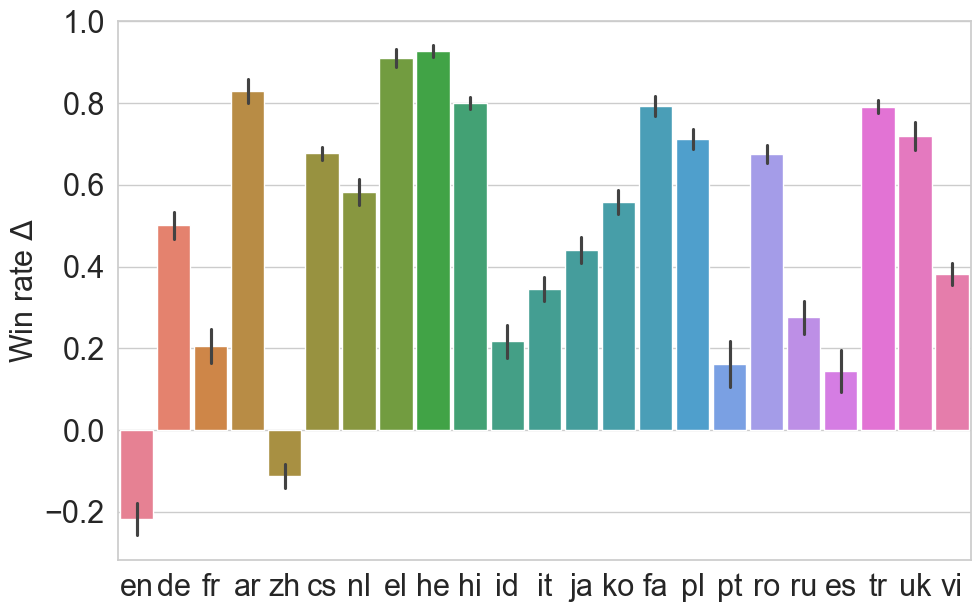}
    \captionof{figure}{Win rates deltas for \textsc{Aya Expanse 8B} vs. \textsc{Qwen2.5 7B} on mArenaHard prompts ($23$ languages), with \textsc{gpt}-4o as a judge. Error bars denote std. dev. across $5$ samples for each prompt.% 
    }
    \label{fig:winrates_by_language}
\end{minipage}
\hfill
    \begin{minipage}[t]{0.4\textwidth}%\vspace{0pt}%
    \centering
     \includegraphics[scale=0.3]
{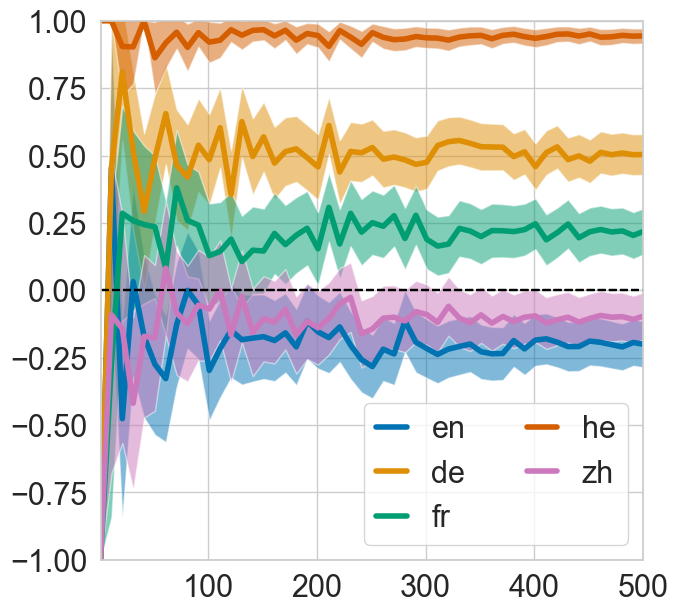}
    \captionof{figure}{Win rate deltas 
    in relation to sample size. Differences are significant when the 95\% confidence interval (shaded) lies above/below zero.}
    \label{fig:sample_size}
\end{minipage}
\end{figure}

% \paragraph{Experiment: Win-rate dependence on sample size} 
\faFlask{} To illustrate the benefits of statistical significance testing, we inspect pairwise comparisons of \textsc{Aya Expanse 8B} and \textsc{Qwen2.5 7B Instruct} on the $500$ prompts of the mArenaHard benchmark for $23$ languages, with \textsc{GPT4}-o as a judge. 
This comparison was previously reported~\citep{dang2024ayaexpansecombiningresearch}, but without considering significance tests or sample sizes.
We compare win rates across languages and compute significance based on 95\% confidence intervals, as recommended by~\citet{miller2024addingerrorbarsevals}.
Experimental details are in~\Cref{app:winrates}. %According to \citet{dang2024ayaexpansecombiningresearch}, Qwen2.5 is one of the closest competitors to Aya Expanse. 

\faSearch{} \textsc{Aya Expanse 8B} wins with an average win rate delta of 0.49 across languages and five runs. 
Individual win rate differences vary widely between languages, from -0.21 (loss) for English to 0.93 for Hebrew (\Cref{fig:winrates_by_language}).
Moreover, \textbf{the significance of the wins is dependent on sample sizes and languages}: ~\Cref{fig:sample_size} illustrates how these win rate differences behave under different sample sizes.
A few Hebrew samples already reveal significant win rate differences, whereas even 300 Chinese samples are insufficient.
This analysis highlights what our current setup cannot reliably determine, namely, whether the models qualitatively differ in Chinese text generation capabilities. For smaller test sets, as in most generative benchmarks~\Cref{sec:statusquo}, such analysis is essential to avoid overconfidence in low-power evaluations.

\recommendation{Test the statistical significance of evaluation results rather than relying on metric differences alone, following task-specific recommendations~\citep{dror-etal-2018-hitchhikers,miller2024addingerrorbarsevals,ackerman2025statisticalmultimetricevaluationvisualization}. Estimate statistical power, particularly when working with small sample sizes. Additionally, consider the magnitude of the effect size to determine whether observed differences are also meaningful in practice.}

\subsection{What Gets Lost In Averages? Aggregating Responsibly}\label{sec:aggregation}

With \mllm{}s, we are modeling multiple languages and tasks at once. 
How we aggregate results thus naturally informs the interpretation of model comparisons. The go-to approach is to report uniformly weighted averages across languages and tasks. This is not necessarily a fair evaluation \---\ due to differences in training distributions \---\ nor is it expressive enough \---\ as outliers (\textit{e.g.}, by unseen languages) can disproportionately affect system rankings~\citep{cohereblog}. Languages and tasks also differ in their expressive power, as seen in~\Cref{sec:statistical_power}. 

In multilingual \mt{}, several aggregation formats were explored beyond reporting plain averages across languages.
For instance, grouping by language resourcedness, \textit{e.g.} to study language-specific routing~\citep{zhang2021share};
by directionality~\citep{zhang-etal-2020-improving}; 
by unseen/seen languages~\citep{aharoni-etal-2019-massively} to isolate zero-shot generalization.
Additionally, WMT offers a constrained track to isolate model improvements from data gains.

\newcommand{\rankbox}[2]{%
    \tikz[baseline=(X.base)] 
    \node[draw=black, 
          fill=#1, 
          rounded corners=3pt, 
          inner sep=1pt, 
          minimum width=12pt, 
          minimum height=12pt, 
          text width=12pt, 
          align=center] (X) {\textbf{#2}};%
}

\newcommand{\rankone}{\rankbox{rankone}{1}}    % Darkest teal
\newcommand{\ranktwo}{\rankbox{ranktwo}{2}}    % Dark teal
\newcommand{\rankthree}{\rankbox{rankthree}{3}}  % Muted teal
\newcommand{\rankfour}{\rankbox{rankfour}{4}}   % Light teal
\newcommand{\rankfive}{\rankbox{rankfive}{5}}   % Very light teal
\newcommand{\ranksix}{\rankbox{ranksix}{6}}   % Very light teal
\newcommand{\rankseven}{\rankbox{rankseven}{7}}   % Very light teal
\newcommand{\rankeight}{\rankbox{rankeight}{8}}   % Very light teal

\begin{table}[t]
    \centering
    \renewcommand{\arraystretch}{1.2} 
    \resizebox{\linewidth}{!}{
    \begin{tabular}{lccc|ccc|ccc|ccc}
     &  \multicolumn{3}{c}{All}  & \multicolumn{3}{c}{High} & \multicolumn{3}{c}{Medium} & \multicolumn{3}{c}{Low} \\
     \midrule
    Models &   Avg & GSM8k & MMLU &  Avg & GSM8k & MMLU &  Avg & GSM8k & MMLU &  Avg & GSM8k & MMLU \\
     \midrule
\textsc{Qwen2-7B} & \rankone & \rankone & \ranktwo & \rankone & \rankone & \ranktwo & \rankone & \rankone & \rankthree & \rankone  & \rankone  & \rankfour \\
\textsc{Mistral-Nemo-Base-12.2B\_2407} & \rankthree & \rankthree & \rankthree & \rankthree& \rankthree & \rankthree & \ranktwo & \ranktwo & \ranktwo & \ranktwo  & \ranktwo & \rankone \\
\textsc{Mixtral-8x7B-v0.1} & \ranktwo & \ranktwo & \rankone  & \ranktwo  & \ranktwo & \rankone& \rankthree & \rankthree & \rankone & \rankfour & \rankfour & \rankthree \\
\textsc{Gemma-7B} & \rankfive & \rankfive & \rankseven &  \rankfive&  \rankfive &  \rankfive & \rankfour & \rankfive  & \rankfour & \rankthree   & \rankthree & \rankfive\\
\textsc{Mistral-NeMo-Minitron-8B-Base} & \rankfour & \rankfour & \rankfour & \rankfour & \rankfour  & \rankfour  & \rankfive & \rankfour & \rankeight & \rankfive & \rankfive & \rankeight \\
    \end{tabular}}
    \caption{Effect of different aggregation strategies on model ranking of top-5 pretrained systems as generated by the European Leaderboard on GSM8k and MMLU datasets. }
    \label{tab:aggregation}
\end{table}

\faFlask{} Table~\ref{tab:aggregation} shows the ranking of the top 5 systems obtained using the \href{https://huggingface.co/spaces/openGPT-X/european-llm-leaderboard}{European Leaderboard} under different configurations: a) by language and b) by task. We categorize languages based on number of speakers into high ($>$ 50M+; en, es, pt, de, fr, it, pl), medium  ($<$ 50 and $>$ 10M; nl, el, hu, sv, cz, ro) and low ($<$ 10M; dk, fi, sk, sl, bg, lt, lv, et) resource.\footnote{\url{https://en.wikipedia.org/wiki/List_of_languages_by_number_of_native_speakers}} 

\faSearch{} Based on average scores, we would conclude that \textsc{Mixtral-8x7B-v0.1} is the second best system after \textsc{Qwen2-7B}, whereas when looking at task-specific aggregates, we find it consistently outperforms \textsc{Qwen2-7B} on \textsc{mmlu}. For medium and low-resource languages, however, its performance for GSM8k drops, leaving the second rank to others. This shows, that \textbf{system rankings can shift based on task and language focus}. Optimal model selection for a specific task and language group can thereby deviate from the average best system.   
%Top system rankings can shift with language- and task-specific evaluation: for instance, \textsc{Mistral-NeMo-Minitron-8B-Base} ranks fourth overall but drops four places on medium and low-resource languages.  

\recommendation{When comparing models across multiple languages, consider differences in language support and aggregate results according to languages being seen by the multilingual models in question. Report task and language-specific scores as a supplement to averages. When you discuss averages, take language coverage into account.}

\begin{figure*}[t]
    \centering
    \includegraphics[width=0.38\linewidth]{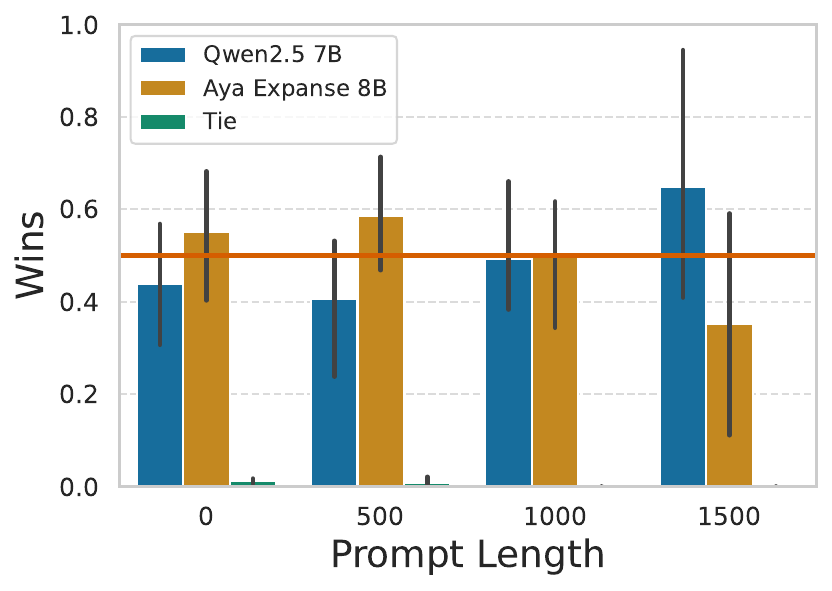}\hspace{2em}
    \includegraphics[width=0.38\linewidth]{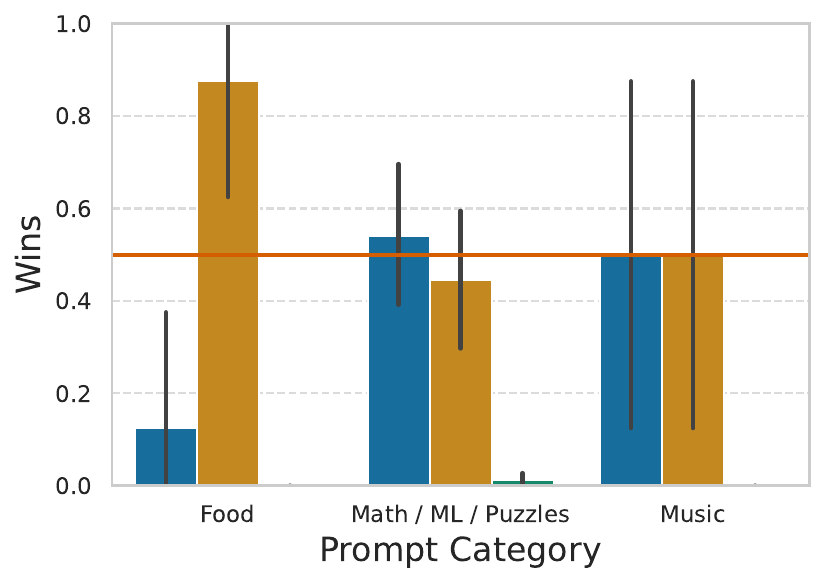}
    \caption{Win-rates for \textsc{Aya Expanse $8$B} vs. \textsc{Qwen$2.5$ 7B} on mArenaHard prompts bucketed by a) prompt length (left) and b) prompt categories (right). }
    \label{fig:win_by_length}
\end{figure*}

\subsection{Where Do Models Differ? Conducting Richer Analyses}\label{sec:analyses}

Aggregate benchmark metrics do not provide insights into what differentiates the outputs of two models—yet identifying these distinctions is often the first step in human preference evaluation. In \mt{}, specialized tools were developed to facilitate pairwise comparisons on specific examples, such as \texttt{MT Compare Eval}~\citep{Klejch2015MTComparEvalGE} and \texttt{compare-mt} \citep{neubig-hu-2018-rapid}. In parallel, there has been a steady effort to create challenge sets and test suites designed to probe particular capabilities and phenomena of \mt{} \citep{stanovsky-etal-2019-evaluating, bawden-sagot-2023-rocs, manakhimova-etal-2024-investigating}.
In contrast, \textsc{llm} evaluations typically rely on user preferences in an arena setting or automatic judges with limited explainability.
Before investing in human evaluation, automatic metrics can already offer insights into quality differences between model outputs across languages. Auxiliary metrics such as diversity scores, length statistics, bias detectors (\textit{e.g.}, toxicity), language confusion statistics, and edit distance can highlight key trends. While these do not fully explain model differences as humans might, they can reveal biases, such as verbosity, that could influence both human and automatic pairwise evaluations.
As \citet{gehrmann2023repairing} put it, there is a ``systemic difference between selecting the best model and characterizing
how good this model really is''.

% \begin{figure*}[!h]
%     \centering
%     \includegraphics[width=0.45\linewidth]{figures/win_by_category.pdf}
%     \caption{Win-rates for \textsc{Aya Expanse $8$B} vs. \textsc{Qwen$2.5$ 7B} on mArenaHard prompts bucketed by prompt categories.}
%     \label{fig:win_by_category}
% \end{figure*}

\faFlask{} We illustrate this by comparing the win rates of \textsc{Aya Expanse 8B} and \textsc{Qwen2.5 7B Instruct} on a subset of languages (en, de, fr, zh) from the mArenaHard benchmark bucketed by a) the prompt length and b) manually annotated prompt category in~\Cref{fig:win_by_length}. 

\faSearch{} The first plot shows a clear trend:  \textbf{\textsc{Qwen2.5 7B Instruct} tends to win on longer prompts, while \textsc{Aya Expanse 8B} performs better on shorter prompts}, suggesting that \textsc{Qwen2.5 7B Instruct} can handle detailed and long queries better. On the other hand, from the second plot, \textsc{Aya Expanse 8B} emerges victorious in all categories except for ``Math / ML / Puzzles'' problems, where \textsc{Qwen2.5 7B Instruct} has a clear advantage. These results provide valuable insights: while the average win rates in Figure~\ref{fig:winrates_by_language} suggest a general preference for \textsc{Aya Expanse 8B}, they obscure \textsc{Qwen2.5 7B Instruct}’s clear advantage on specific prompt types. Such findings can guide targeted test set design, inform human evaluation sampling, and steer future model development.

\recommendation{Complement automatic metric analyses with qualitative error analysis to better understand systematic patterns. Use visualization and systematic category breakdowns to contextualize metric results, ensuring that observed differences align with meaningful distinctions rather than incidental artifacts.}

\begin{figure}[t]
  \centering
  \makebox[0.9\textwidth]{
    \begin{minipage}[t]{0.42\textwidth}
      \centering
      \vspace{-2.45cm}
      \resizebox{\textwidth}{!}{
        \begin{tabular}{lccc}
          \toprule
           & EN 1 & EN 4 & EN 6 \\
          \midrule
          Aya Expanse 8B & 46.0\% & 48.9\% & 42.4\% \\
          Llama 3.1 8B & \textbf{50.4\%} & 48.2\% & \textbf{51.1\%} \\
          Gemma 2 9B & 48.2\% & \textbf{49.6\%} & \textbf{51.1\%} \\
          \midrule
          Score range & 4.4\% & 1.4\% & 8.7\% \\
          \bottomrule
        \end{tabular}
      }
      \phantomcaption \label{tab:instruction_wording}
    \end{minipage}
    \hspace{1em}
    \begin{minipage}[t]{0.42\textwidth}
      \centering
      \includegraphics[width=\linewidth]{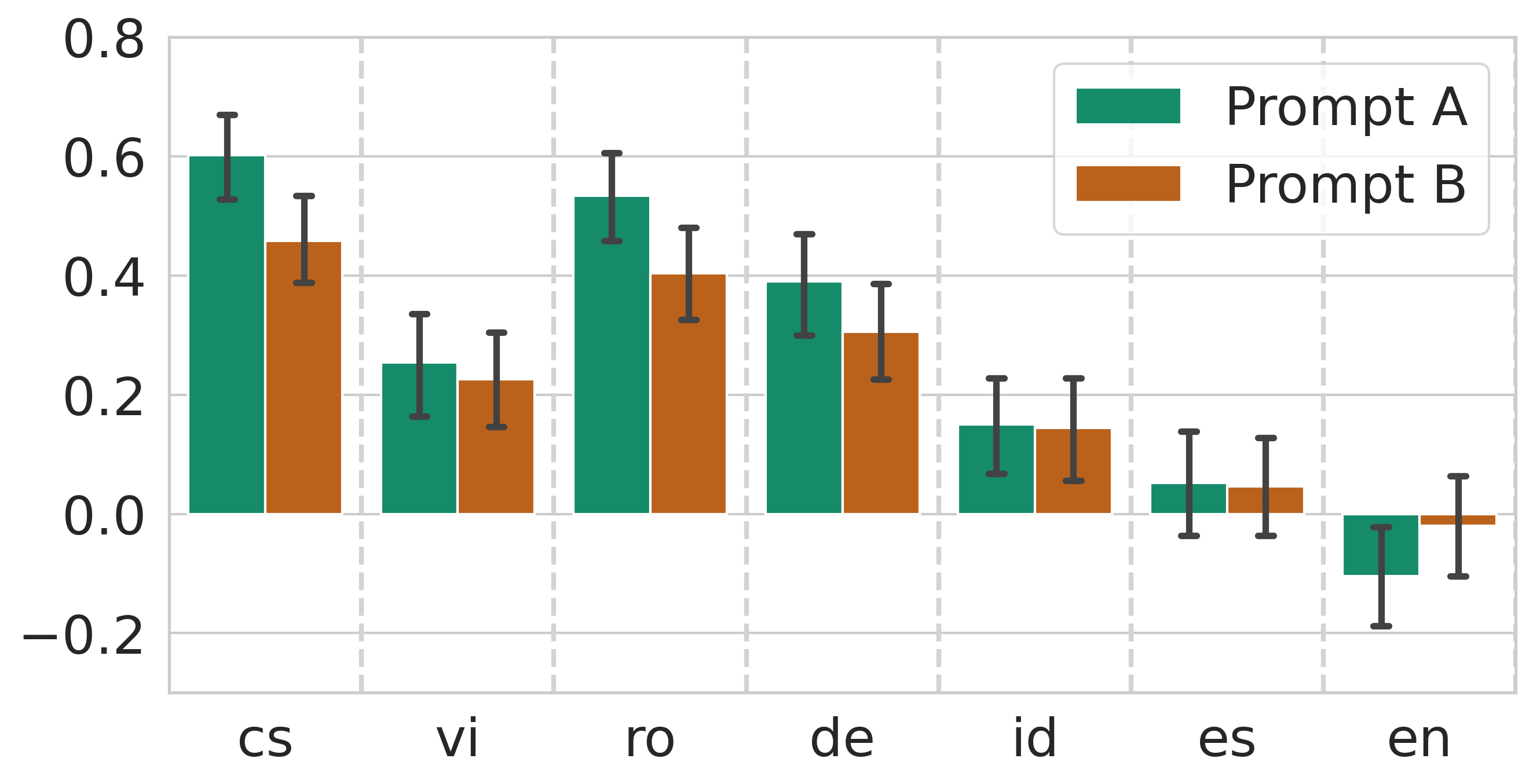}
      \phantomcaption \label{fig:prompt_formulation_judge_bias}
    \end{minipage}
  }\vspace{-1.5em} % Tighten space between top and bottom rows
  \makebox[0.9\textwidth]{
    \begin{minipage}[t]{0.42\textwidth}
      \centering
      \includegraphics[width=\linewidth]{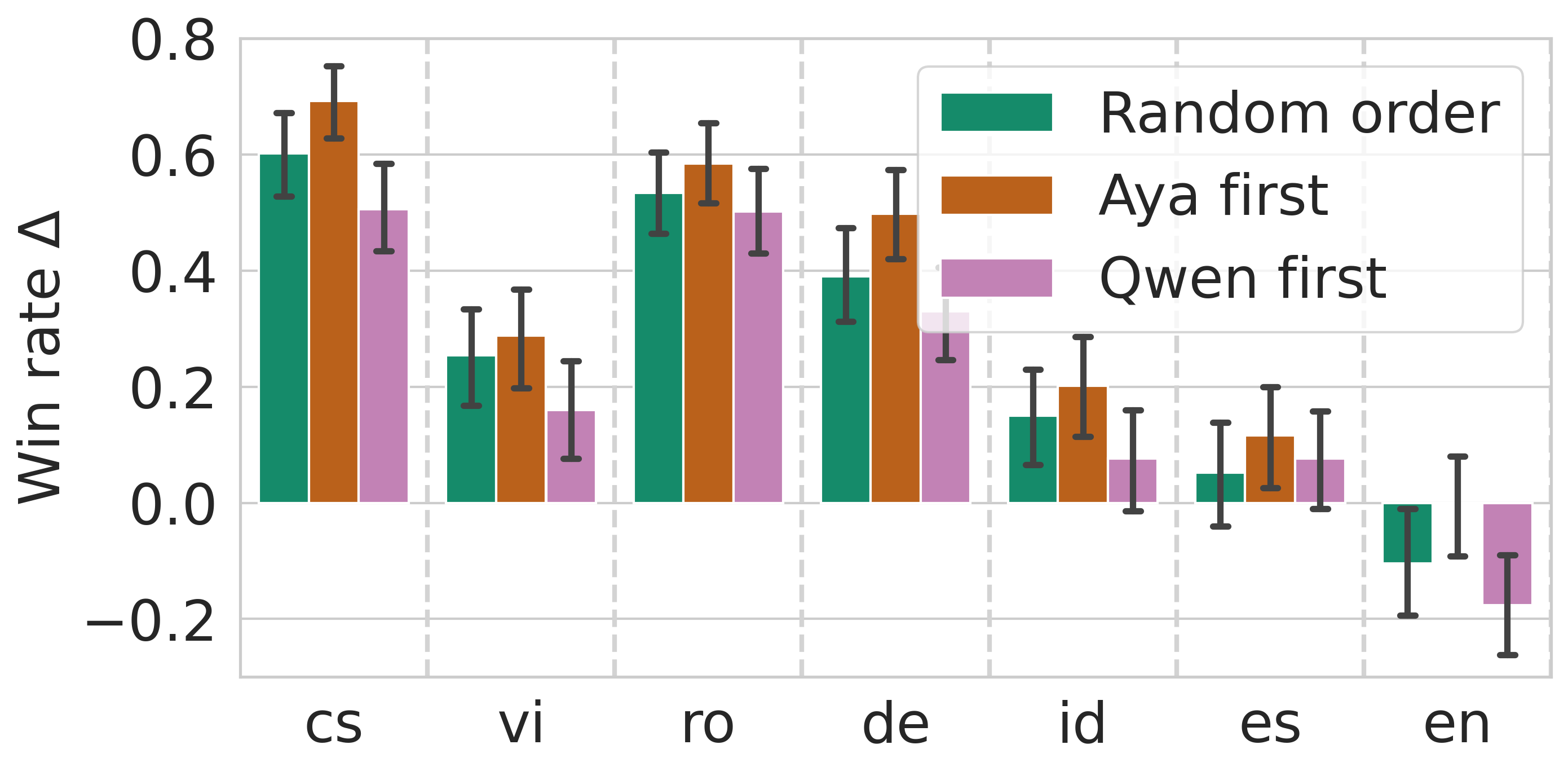}
      \phantomcaption \label{fig:positional_judge_bias}
    \end{minipage}
    \hspace{1em}
    \begin{minipage}[t]{0.42\textwidth}
      \centering
      \includegraphics[width=\linewidth]{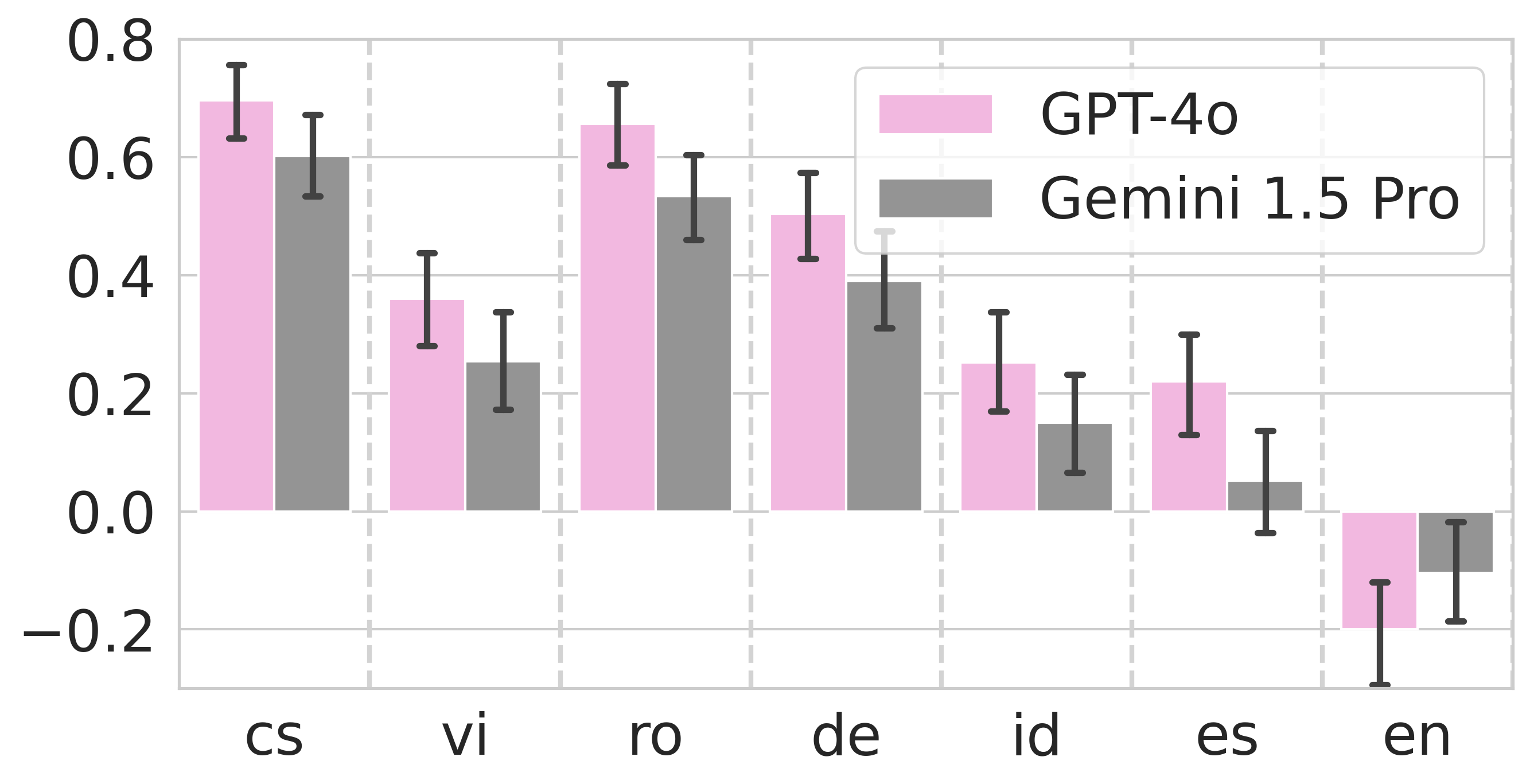}
      \phantomcaption \label{fig:judge_bias}
    \end{minipage}
  }
  \vspace{-0.85em} % Tighten space before caption
  \caption{Accuracy on German MCQA (Include 44) with three instruction variants (top-left). Win-rates of Aya Expanse 8B vs. Qwen 2.5 7B on mArenaHard prompts showing prompt (top-right), positional (bottom-left), and judge bias (bottom-right).}
  \label{fig:multi_panel_bias}
\end{figure}

\subsection{What Do We Need to Share? Advancing Reproducibility Through Transparency}\label{sec:reproducibility}

Reproducing evaluation results in the \textsc{llm} era has become increasingly challenging, if not impossible~\citep{vaugrante2024loomingreplicationcrisisevaluating}. 
Not only are many evaluations stochastic, but they are also dependent on configurations that are rarely fully disclosed, such as preambles or system prompts, task formatting, decoding strategies, temperature, or answer parsing. 
A similar challenge arose in \mt{}, where even a straightforward metric like \textsc{bleu} was implemented differently across frameworks, leading to discrepancies in reported scores. The introduction of Sacre\textsc{bleu}~\citep{post-2018-call} marked a turning point by standardizing the evaluation pipeline into a single toolkit, with each evaluation assigned a unique signature containing all relevant parameters, ensuring comparability across papers.
Efforts like \texttt{simple-evals}\footnote{\small \url{https://github.com/openai/simple-evals}} and the LM Evaluation Harness~\citep{eval-harness} aim to standardize task formulations and output parsing for \textsc{llm} evaluations. 
However, full transparency requires \emph{open evaluation releases} that contain publicly available code with exact versioning (\textit{e.g.}, commit hashes), full release of all prompts (including instruction text, exact wording, punctuation, and formatting), and disclosure of task formulations in each language.\footnote{As an example, we release the pairwise evaluation artifacts from this paper: \url{https://huggingface.co/datasets/CohereLabs/deja-vu-pairwise-evals}.}
For example, \citet{Briakou2024OnTI} 
demonstrate that minor variations in prompt wording for translation tasks can lead to drastically different outcomes, such as models refusing to translate or producing overly verbose responses. 
Such findings are enabled by the practice of releasing model outputs, championed in the annual WMT shared task competitions~\citep{ws-2006-statistical}, and has kindled metrics and meta-evaluation research by allowing retroactive comparisons and enabling longitudinal studies~\citep{graham-etal-2014-machine}.

\faFlask{} We illustrate configuration's impact on accuracy results
for German \textsc{mcqa} (\textsc{include} $44$~\citep{romanou2024include}) ($3$ prompts) and mArenaHard \textsc{llm}-as-a-judge win rates varying a) the prompt, b) the compared systems' order, and c) the judge (\textsc{gpt}-4o vs Gemini 1.5Pro).

\faSearch{} ~\Cref{fig:multi_panel_bias} shows that system accuracy on \textsc{mmlu}-like evaluations changes significantly with different instruction wordings, undermining the robustness of benchmarking~\citep{alzahrani-etal-2024-benchmarks}.
%We observe similar sensitivity in our analysis of the \textsc{include} dataset~\citep{romanou2024include}, where system accuracy on \textsc{mmlu}-like evaluations changes significantly with different instruction wordings (see~\Cref{fig:multi_panel_bias}). 
%
The use of \textsc{llm}s as judges further complicates reproducibility. Variability in model choice, decoding strategies (\Cref{app:sampling}), various biases~\citep{ye2024justice,shimabucoro-etal-2024-llm}, and prompt phrasing adds layers of complexity. Figure~\ref{fig:multi_panel_bias} illustrates how evaluations can be manipulated through positional biases (system presentation order) and prompt formulation differences, yielding significantly divergent outcomes.
Finally, \textsc{llm} version obsolescence prevents reliable comparisons of results across papers and over time. Evaluations are often non-transitive~\citep{xu2025investigatingnontransitivityllmasajudge}, making optimization a moving target.

\recommendation{Use standardized pipelines, publish the exact prompt wording, and release the evaluation code, model outputs, and evaluation scores with versioning.}

\begin{figure}[t!]
\begin{minipage}{\textwidth}
    
    \begin{minipage}[b]{0.7\textwidth}\vspace{0pt}%
        \centering
    \includegraphics[width=\linewidth]{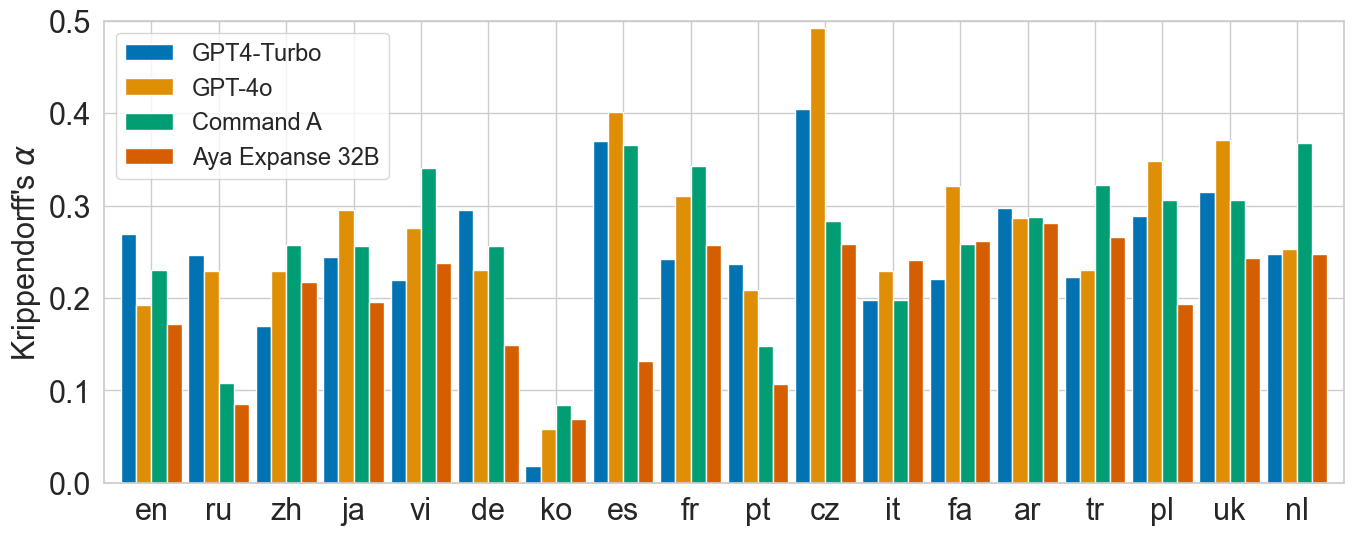}
    \captionof{figure}{LLM-as-a-judge agreement with humans on arena.}
            \label{fig:arena_judges_pairwise}

    \end{minipage}\hfill
    \begin{minipage}[b]{0.25\textwidth}\vspace{0pt}%
    \centering
    \resizebox{\linewidth}{!}{%
    \begin{tabular}{lc}
    \toprule
      %  \textbf{LLM Judge} & \\
      %           \midrule

        & \textbf{$\alpha$}  \\
        GPT-4-Turbo  &  0.25  \\
GPT-4o    &     \textbf{0.28}  \\
%Command-R+ &   0.19 \\
Command A & 0.26\\
Aya Expanse 32B   &   0.20 \\
\midrule
& \# Best \\
GPT-4-Turbo  & 5  \\
GPT-4o    &    \textbf{6}   \\
%Command-R+ &  1 \\
Command A & \textbf{6}\\
Aya Expanse 32B   &  1  \\
    \bottomrule
    \end{tabular}%
    }
    \captionof{table}{Summary across languages.} 
    \label{tab:avg_agreement}
    \end{minipage}%
\end{minipage}
\end{figure}

\section{Evaluating mLLM Evaluation: Towards Meta Evaluation}\label{sec:metaevaluation}

The field of \mt{} has been consistently involved in meta-evaluation of machine translation evaluation methods for the last twenty years \citep{callison-burch-etal-2007-meta}. This process, whether implicit or explicit, has driven progress in \mt{} systems by redefining which metrics best correlate with human judgments throughout various milestones of model improvements.  Given this progress in \mt{} evaluation, a natural question arises: why has not similar progress been observed in multilingual evaluation? To better understand this disparity, we revisit the prerequisites of meta-evaluation. 
Meta-evaluation fundamentally requires three components: system outputs, human judgments (of those outputs), and automatic evaluations of those same outputs. In the following sections, we identify the missing components in the multilingual setting and outline the steps necessary to overcome these challenges.

%%%%%%%%%%%%%%%%%%%%%%%%%%%%%%%%%%%%%%%%%%%%%%%%%%%%%%%%%%%%%%%%%%%%%%%%
\subsection{Need for (More) Metrics: Beyond One-size-fits All}\label{sec:need_for_metrics}

\textbf{Lesson from \mt{}}
While \llm{}s-as-a-judge offers the convenience of using a single model for multilingual assessments, \mt{} meta-evaluation shows there is no universally best metric~\citep{marie-etal-2021-scientific,anugraha-etal-2024-metametrics}. Central to this issue was the widespread use of pretrained language models as backbones for developing learned metrics~\citep{lo-2020-extended}, with varying degrees of language representation coverage. Although learned metrics often outperform string-based ones \---\ like \textsc{bleu} \citep{papineni-etal-2002-bleu} \---\ the choice between a string-based and a learned metric is heavily language and domain dependent. This has led to the informal convention of employing different types of metrics for targeted evaluations and reporting multiple metrics. % to support findings.
%\end{mtlesson}

%\smallskip\noindent
\textbf{Application to multilingual evaluation}
%\faFlask{}
\citet{zheng2023judging} used human evaluations from Chatbot Arena~\citep{chiang2024chatbotarenaopenplatform} to study the reliability of \llm{} judges. %While some non-English prompts were included in the data, the analysis did not analyze performance across languages separately. 
We extend their predominantly English analysis to non-English ``battles" (pairwise comparisons) from the released sample of battles (\href{https://huggingface.co/datasets/lmarena-ai/arena-human-preference-100k}{\texttt{lmarena-ai/arena-human-preference-100k}}), focusing on the 18 languages with $>$200 prompts.  We score a subset of 200 generation pairs for each language with two open and two closed \mllm{}s as judges.  We measure agreement with human preferences with Krippendorff's $\alpha$ (interval measurements),
shown in
~\Cref{fig:arena_judges_pairwise}.
Overall agreement, and the choice of the best judge varies across languages.
Just like in \mt{}, the optimal choice of an \llm{} judge (i.e., metric) for multilingual evaluation is language dependent (\Cref{tab:avg_agreement}).

%%%%%%%%%%%%%%%%%%%%%%%%%%%%%%%%%%%%%%%%%%%%%%%%%%%%%%%%%%%%%%%%%%%%%%%%
% Human Evaluation

\subsection{Need for Nuanced Human Evaluation: Towards Richer Assessments}\label{sec:need_for_human}

\textbf{Lesson from \mt{}} 
Human evaluation of translation quality is a multifaceted challenge, rooted in the fundamental questions of \textit{what} to measure and \textit{how} to elicit accurate, consistent human assessments. These questions have long been a central focus in machine translation.
Below, we highlight key areas and insights from this extensive body of work.

Detailed work has explored different \textbf{\textit{evaluation protocols}}~\citep{vilar-etal-2007-human, graham-etal-2013-continuous, graham-etal-2014-machine} and \textbf{\textit{quality dimensions}}, including fluency versus adequacy~\citep{koehn-monz-2006-manual, bojar-etal-2016-findings}. 
The trend then went from monolithic assessment scores of pair-wise assessments towards more \textbf{\textit{fine-grained protocols}}, \textit{e.g.}  %These methods include
highlighting and annotating errors using established taxonomies, such as the Multidimensional Quality Metrics (\textsc{mqm}) framework~\citep{burchardt-2013-multidimensional, freitag-etal-2021-experts}.
These taxonomies are getting refined to balance cognitive load and annotation effectiveness~\citep{Ge2024ESAAA}, and adapted for non-professional annotators~\citep{Graham2015CanMT, castilho-etal-2017-comparative, wang-etal-2024-afrimte}. 
Annotation efforts have also expanded to target specific use cases, such as \textbf{\textit{critical error}} detection~\citep{specia-etal-2021-findings, zerva-etal-2022-findings}, detection of errors grounded in high-stake scenarios~\citep{mehandru-etal-2023-physician}, and to contextualize evaluations within \textbf{\textit{user-centric frameworks}} \citep{briakou-etal-2023-explaining, Savoldi2025TranslationIT}. 

%\smallskip\noindent
\textbf{Application to multilingual evaluation} 
Chatbot Arena, where users compare two models in a chat and choose a winner, are the primary source of public human \mllm{} evaluations. The maintainers periodically releases data,\footnote{\url{https://github.com/lm-sys/FastChat/blob/main/docs/dataset_release.md}} forming the largest public collections of multilingual human preferences. In the data from 2024, 27--43\% of battles per language end in a tie  (analysis in~\Cref{app:chatbot_arena}), suggesting that human evaluation in the  pairwise arena format lacks sensitivity to fine-grained differences or inherently includes significant uncertainty. The scarcity of publicly available multilingual preference data limits research in this area.
Emerging efforts such as MM-EVAL~\citep{son2024mm} introduce multilingual meta-evaluation benchmarks, highlighting that LLMs-as-a-judge often lack fairness and consistency across languages.

%%%%%%%%%%%%%%%%%%%%%%%%%%%%%%%%%%%%%%%%%%%%%%%%%%%%%%%%%%%%%%%%%%%%%%%%

\subsection{Need for meta-evaluation research: Closing the Loop on Evaluation}

\textbf{Lesson from \mt{}} 
Meta-evaluation research has been formally conducted within the \textsc{wmt} Metrics shared task since 2007~\citep{callison-burch-etal-2007-meta}. This research aims to improve \mt{} evaluation by identifying best performing metrics, addressing weaknesses in correlation-based metrics, \textit{e.g.}, proper handling of ties~\citep{deutsch-etal-2023-ties}, meta-evaluation techniques \citep{kocmi-etal-2021-ship, thompson-etal-2024-improving}, and challenges of conducting reliable human evaluations, such as ensuring replicable human evaluations~\citep{riley-etal-2024-finding} and studying inter-annotator agreement~\citep{popovic-2021-agree, popovic-belz-2022-reporting}. 

%\smallskip\noindent
\textbf{Application to multilingual evaluation}
Evaluation solely based on correlation with human pairwise preferences on prompt-level comparisons is not sustainable, as human agreement decreases with shrinking quality differences between the contrasted systems~\citep{zheng2023judging}  \---\ a finding that was also observed in \mllm{} development~\citep{ustun-etal-2024-aya}. 
We get a glimpse of this loss of signal in Chatbot arena battles: from 2023 to 2024, the ratio 
of ties has grown significantly, from an average of 29\% to 40\% across the six most dominant languages (\Cref{app:chatbot_arena}).
For long-term progress in \mllm{} modeling and evaluation, we need to iterate the meta-evaluation loop. As a first step, we need to answer the questions which differences between models matter to humans and how to capture these. Then, we can adapt automatic metrics accordingly. Finally, we can   
measure modeling progress automatically and reliably, which results in models with enhanced qualities, bringing us back to the first step.

\section{Conclusion}
\textsc{MT} has long grappled with the complexities of multilingual generative evaluations, from constructing datasets to benchmarking evaluation metrics. We demonstrated that established practices from \textsc{mt} can enhance the understanding and reliability of comparisons among \mllm{}s, and outlined which elements are necessary for establishing meta-evaluations. Our recommendations are distilled into a practical checklist (\Cref{app:checklist}).

\textbf{Limitations} Our experiments are focused on open-ended generative tasks, which could be still considered more ambiguous than typical \mt{} tasks, since \mt{} evaluation criteria are more defined with respect to a reference generation. 
Therefore, not all approaches for \mt{} evaluation might transfer equally well. 
However, with more advances in quality, \mllm{} evaluations are trending towards more tightly defined benchmarks that require in-depth expert knowledge (e.g. coding, math), which brings these tasks closer to the conditions of \mt{} evaluations.
Furthermore, recommendations and best practices from other sub-fields of NLP that are now sharing multilingual benchmarks should also be considered, see for example the recommendations by~\citet{gehrmann2023repairing} for evaluating text generation, or those by \citet{iskender-etal-2021-reliability} for human evaluation of text summarization.

\textbf{Outlook} Since \mllm{}s are now also competing with non-\textsc{llm} \mt{} models~\citep{kocmi-etal-2023-findings,kocmi-etal-2024-findings,zhu-etal-2024-multilingual}, and \mt{} benchmarks have become established evaluation tasks for \mllm{}s~\citep{zhu2024multilinguallargelanguagemodels}, the sharing of knowledge and insights across both disciplines becomes even more important to drive meaningful progress. 
Our checklist with practical recommendations for \mllm{} evaluations is the first step towards the aim of bringing research communities closer.

\section*{Acknowledgments}
We thank the anonymous COLM reviewers and our colleagues for their helpful feedback on the paper: Shivalika Singh, John Dang, Colin Cherry.

\bibliography{anthology, custom}

\appendix
\clearpage

\section{Inspected Multilingual Generative Benchmarks}
\begin{table*}[]
    \centering
    \resizebox{\textwidth}{!}{%
    \begin{tabular}{lllp{5cm}lll}
    \toprule
        \textbf{Benchmark }& \textbf{Test Size }& \textbf{Metric(s)} & \textbf{Source} & \textbf{\#Langs} & \textbf{Translated?}\\
    \midrule
        \textit{Translation} & \\
        Flores-200 \citep{nllb2022} & $\approx$1k & Comet-22, ChrF++, spBLEU & Wikinews, Wikijunior, Wikivoyage & 200 & Human\\
        % Flores-In^{*} & \citep{singh-etal-2024-indicgenbench} & 1012 & ChrF & translated Flores & 29 & Human \\ removed by Kocmi - it is only extension of Flores
        NTREX-128 \citep{federmann-etal-2022-ntrex} & $\approx$2k & Comet-22, ChrF++, spBLEU &  News from 2019 & 128 & Human \\
        WMT General MT \citep[inter alia]{kocmi-etal-2024-findings} & $\approx$2000 &  Comet-22 &  News, literary, e-commerce, social, speech& $\approx$11 & Human\\ 
        %WMT-24 General MT & \citep{kocmi-etal-2024-findings}  &  $\approx$2000 &  Comet-22 &  News, literary, social , speech& 9 & Human\\
        %WMT-23 General MT & \citep{kocmi-etal-2023-findings} &  $\approx$2000& Comet-22 & News, e-commerce, manuals, social, speech & 9 & Human \\
        %WMT-22 General MT & \citep{kocmi-etal-2022-findings} &  $\approx$2000& Comet-22 & News, e-commerce, conversation, social & 11 & Human \\
        
       %  WMT-14 & tbd & tbd & BLEU\footnote{All WMT evaluations originally include human evaluations, that may or may not correlate with automatic metrics. For ease of use, automatic metrics have been adopted for \mllm{} evaluation.} & tbd & tbd & \\
       % WMT-16 & \citep{bojar-etal-2016-findings} & 3000 & BLEU& Online news sources & 6 & \\
       % WMT-20 & tbd & tbd & BLEU & Online news sources & tbd & \\ 
       MAFAND-MT  \citep{adelani-etal-2022-thousand} & 1000 & ChrF & Online news sources & 21 &  \\ % no model release benchmarked on it, https://huggingface.co/datasets/masakhane/mafand
        \midrule
        
        \textit{Summarization} \\
        XLSum   \citep{hasan-etal-2021-xl} &   500--11k & ROUGE & BBC News& 45 & -\\
         CrossSum-In  \citep{singh-etal-2024-indicgenbench} & 500 & ChrF & translated XLSum & 29 & Human \\ % 20 new
        \midrule
        \textit{Math} \\
        MGSM \citep{shi2023language-mgsm} & 250 & Accuracy & GSM8K & 10 & Human\\
        AfriMGSM  \citep{adelani2024irokobenchnewbenchmarkafrican} & 250 & Accuracy & translated MGSM & 16 & Human \\ 
        \midrule
        \textit{Open-ended generation} \\
        MTG  \citep{chen-etal-2022-mtg} & 3000 & derived from ROUGE & translated English tasks with human post-edits & 5 & Google Translate API\\
        OMGEval  \citep{liu2024omgevalopenmultilingualgenerative} & 804 & win-rate & selected prompts from AlpacaEval, translated, localized, verified & 5 & GPT-4 \\
        mArenaHard  \citep{dang2024ayaexpansecombiningresearch} &  500 & win-rate & LMARENA prompts & 23 & Google Translate API\\ 
        Dolly translated  \citep{singh-etal-2024-aya} &   200 & win-rate & mixed prompts from Databricks employees & 101 & NLLB\footnote{A subset of 6 languages is available with post-edited translations.}\\
        Aya human-annotated  \citep{singh-etal-2024-aya} &   250 & win-rate & community-sourced Aya dataset & 7 & - \\
        PolyWrite  \citep{ji2024emma500enhancingmassivelymultilingual}&  $\approx$155\footnote{Some are filtered out for certain languages based on automatic quality assessment via backtranslation.} & self-BLEU & Writing tasks, generated by ChatGPT & 240 & Google Translate API \\
        MultiQ  \citep{holtermann-etal-2024-evaluating} & 200 & LLM-judged accuracy & selected from LMSYS and GPT-4 generated questions & 137 & Google Translate API \\ % https://huggingface.co/datasets/caro-holt/MultiQ
        \midrule
        \textit{Chat} \\
        %Multilingual MTBench & \citep{ming2024marcollmbridginglanguagesmassive} & 80 & win-rate & Multi-turn chat questions  &  28 & ? \\ % is not open
       SeaBench  \citep{liu-etal-2025-seaexam} & 300 & LLM score against reference & human written and localized &  3 & - \\
       Sea-MTBench  \citep{sea_lion_2024} & 58 & LLM score against baseline & translated MTBench & 6 & Human \\ % https://huggingface.co/datasets/aisingapore/multiturn_chat-mtbench 
       %SeaWildBench & \citep{} &  & LLM score against baseline & translated WildBench & 8 & GPT-4o0806 \\ % not public
       \midrule
       \textit{Format Following} \\
       SEA-IFEval  \citep{sea_lion_2024} & 105 & Accuracy & translated IFEval & 6 & Human \\ % https://huggingface.co/datasets/aisingapore/instruction_following-ifeval
        MIFEval \citep{zhang2024pmmevalparallelmultilingualmultitask} & 96 & Accuracy & translated and post-edited, localized, filtered IFEval & 10 & unspecified LLM \\
       MultiIF  \citep{he2024multiifbenchmarkingllmsmultiturn} & 454--909 & Accuracy & translated and localized IFEval, verified, and expanded with additional turns & 7 & Llama 3.1 405B\\
    \bottomrule
    \end{tabular}%
    }
    \caption{Public generative benchmarks for downstream text-based evaluation of multilingual LLMs. Note that WMT annually releases benchmarks for varying languages and domains that we summarize here under a single item. ``Test size'' counts the number of prompts in the test split per language.
    }
    \label{tab:eval_benchmarks_full}
\end{table*}
\Cref{tab:eval_benchmarks_full} gives an overview of the multilingual generative benchmarks that we inspected for this paper. \Cref{tab:eval_benchmarks} summarizes these more concisely.

\section{Model Release Benchmarks}\label{app:model_release_benchmarks}
\begin{table}[htbp]
\centering
    \resizebox{\textwidth}{!}{%
    \begin{tabular}{lllp{8cm}}
    \toprule
    \textbf{Rank} & \textbf{Benchmark} & \textbf{Paper} & \textbf{Model Releases (Benchmarked/Supported Languages)}\\
    \midrule
        1 & Flores-200 & \citep{nllb2022}& Aya101 (99/101), Aya Expanse (22/23), Qwen2 (?/$\approx$30), EMMA (199/546), EuroLLM (34/35), PangeaLLM (11/39), SeaLLM (12/12), SEA-LION (4/13), Salamandra (3/35), Babel (25/25), Sailor2 (15/15)\\
        \midrule
        2 & MGSM & \citep{shi2023language-mgsm} & Aya Expanse (7/23), Llama3 (7/8), Qwen2 (10?/$\approx$30), EMMA (10/546), PangeaLLM (10?/39), SeaLLM (6/12), Salamandra (5/35), Babel (10?/25)\\
        \midrule
       3 & XLSum & \citep{hasan-etal-2021-xl} & Aya101 (45/101), EMMA (44/546), FuxiTranyu (15/43), SEA-LION (4/13), Salamandra (2/35)\\
       \midrule
       \multirow{2}{*}{4}  & WMT & \citep[inter alia]{kocmi-etal-2024-findings} & EuroLLM (16/35), FuyiTranyu (3/43), PolyLM (4/8+) \\
         & Dolly translated &  \citep{singh-etal-2024-aya} & Aya101 (3/101), Aya Expanse (23/23), EMMA (119/546) \\
         \midrule
       \multirow{7}{*}{5}  & mArenaHard & \citep{dang2024ayaexpansecombiningresearch} & Aya Expanse (23/23) \\
         & Aya human-translated & \citep{singh-etal-2024-aya} & Aya101 (5/101) \\
         & PolyWrite & \citep{ji2024emma500enhancingmassivelymultilingual} & EMMA (240/546) \\
         & SeaBench & \citep{zhang2024seallms3openfoundation} & SeaLLM (3/12) \\
         & SeaMTBench & \citep{tjhi-etal-2023-sea} & SEA-LION (6/13) \\
         & SEA-IFEval & \citep{tjhi-etal-2023-sea} & SEA-LION (6/13) \\
         & MTG & \citep{chen-etal-2022-mtg}  & PolyLM (5/8) \\
        \bottomrule
    \end{tabular}%
    }
    \caption{We rank open benchmarks from Table~\ref{tab:eval_benchmarks} on their popularity in model release reports. 
    For each model we indicate in how many of its supported languages the model is evaluated.
    For WMT General Benchmarks, we report the union of all subsets.}
    \label{tab:eval_benchmark_use}
\end{table}

%  Aya101 (/101), Aya Expanse (/23), Llama3 (/8), Qwen2 (/\approx30), EMMA (/546), EuroLLM (/35), PangeaLLM (/39), FuxiTranyu (/43), SeaLLM (/12), SEA-LION (/13), PolyLM (/8+), Salamandra (/35), Babel (/25), Sailor2 (/15)

Table~\ref{tab:eval_benchmark_use} indicates which of these benchmarks were included in recent open (explicitly) multilingual model releases,\footnote{Other models such as Gemma2~\citep{gemmateam2024gemma2improvingopen} might have multilingual capabilities but are not explicitly stating that they do.}  including 
Aya-101~\citep{ustun-etal-2024-aya}, 
Aya Expanse~\citep{dang2024ayaexpansecombiningresearch}, 
Llama3~\citep{grattafiori2024llama3herdmodels}, 
Qwen2~\citep{yang2024qwen2technicalreport}, 
EMMA-500~\citep{ji2024emma500enhancingmassivelymultilingual} (base model), 
EuroLLM~\citep{martins2024eurollmmultilinguallanguagemodels}, 
PangeaLLM~\citep{yue2024pangeafullyopenmultilingual} (multi-modal), 
FuxiTranyu~\citep{sun-etal-2024-fuxitranyu}, 
PolyLM~\citep{wei2023polylmopensourcepolyglot},
% MarcoLLM~\citep{ming2024marcollmbridginglanguagesmassive}. - is not open :(
SeaLLMs~\citep{zhang2024seallms3openfoundation}, 
SEA-LION~\citep{tjhi-etal-2023-sea}, 
Salamandra~\citep{salamandra}, 
Babel~\citep{zhao2025babelopenmultilinguallarge},
Sailor2~\citep{sailor2report}.\footnote{We excluded models that do not report any generative evaluations, such as Ministral and Mixtral, or those that are only in-house (Qwen). 
Models might additionally have been benchmarked by external parties or competing model releases (\textit{e.g.} EuroLLM benchmarks Gemma on translation tasks).}

\section{Multilingual Leaderboards}\label{sec:leaderboards}
\begin{table*}
\centering
    \resizebox{\textwidth}{!}{%

    \begin{tabular}{lclll}
    \toprule
    & \# Languages & Language Focus &  Evaluated Open \mllm{}s & Focus Language(s) LLM win?\\
    \midrule
       \href{https://huggingface.co/spaces/openGPT-X/european-llm-leaderboard}{European LLM Leaderboard}  & 21 & European  & Llama3, EuroLLM, Qwen2, Aya23 & no\\
%\href{https://huggingface.co/spaces/uonlp/open_multilingual_llm_leaderboard}{Open Multilingual LLM Leaderboard}  & 29 & - & \textcolor{gray}{Llama, BLOOM} & \\ % only two models -> excluded
\href{https://huggingface.co/spaces/taresco/open_african_languages_eval_leaderboard}{African Languages LLM Eval Leaderboard} & 18 & African & Llama3, Aya101 & no\\ % subteasks have various languages, not filtered by language
       \href{https://leaderboard.sea-lion.ai/}{SEA HELM} & 4 & South-East Asian  & Llama3, Qwen2, SeaLLMs, SEA-LION, Aya Expanse, Aya23 & yes \\
\href{https://huggingface.co/spaces/Cognitive-Lab/indic_llm_leaderboard}{Indic LLM Leaderboard} & 7 & Indic & Llama3 & no \\
\href{https://huggingface.co/spaces/llm-jp/open-japanese-llm-leaderboard}{Open Japanese LLM Leaderboard} & 1 & Japanese & Llama3, Qwen2, Aya Expanse & yes \\
\href{https://huggingface.co/spaces/upstage/open-ko-llm-leaderboard}{Open Ko-LLM Leaderboard} & 1 & Korean & Qwen2, SeaLLM & yes\\ % sometimes inactive?
\href{https://huggingface.co/spaces/PartAI/open-persian-llm-leaderboard}{Open Persian Leaderboard} & 1 & Persian & Qwen2, Aya Expanse, Llama3 & no\\
\href{https://huggingface.co/spaces/eduagarcia/open_pt_llm_leaderboard}{Open Portuguese Leaderboard} & 1 & Portuguese & Qwen2, Llama3, Aya Expanse, Aya23, SeaLLM & yes \\
\href{https://huggingface.co/spaces/BAAI/open_cn_llm_leaderboard}{Open Chinese Leaderboard} & 1 & Chinese & Qwen2, Llama3, SeaLLM & yes \\
\href{https://huggingface.co/spaces/OALL/Open-Arabic-LLM-Leaderboard}{Open Arabic Leaderboard} & 1 & Arabic & Qwen2, Llama3, Aya Expanse, SeaLLM, Aya23, EuroLLM & yes \\
\href{https://huggingface.co/spaces/CIIRC-NLP/czechbench_leaderboard}{CzechBench Leaderboard} & 1 & Czech & Llama3 & no \\
\href{https://huggingface.co/spaces/hebrew-llm-leaderboard/leaderboard}{Hebrew LLM Leaderboard} & 1 & Hebrew & SeaLLM, Aya Expanse, Qwen2, Aya23, Llama3 & yes\\

\href{https://huggingface.co/spaces/speakleash/open_pl_llm_leaderboard}{Open PL LLM Leaderboard} & 1 & Polish & Llama3, Qwen2, Aya Expanse, EuroLLM, Aya 23 & yes\\
\href{https://huggingface.co/spaces/malhajar/OpenLLMTurkishLeaderboard_v0.2}{OpenLLM Turkish Leaderboard} & 1 & Turkish & Aya Expanse, Aya 23, EuroLLM, Llama3, Qwen2 & yes\\
%\href{https://huggingface.co/spaces/Vikhrmodels/arenahardlb}{Ru Arena General} & 1 & Russian & \\
\href{https://huggingface.co/spaces/le-leadboard/OpenLLMFrenchLeaderboard}{Open LLM French Leaderboard} & 1 & French & Qwen2, Llama3, EuroLLM & no \\
    \bottomrule
    \end{tabular} %
    }
    \caption{Non-English leaderboards evaluating multilingual models with their focus languages and evaluated open \mllm{}s from Table~\ref{tab:eval_benchmark_use}. Based on the average ranking on the respective leaderboards, we measure if LLMs for the respective focus languages win over the more massively multilingual ones, restricted to models below 13B parameters. For leaderboards that involve multiple languages, we aggregate wins via majority votes. This table reflects the state of 10 February 2025, 7 March 2025 for  the French leaderboard. 
    }
    \label{tab:leaderboards}
\end{table*}

Table~\ref{tab:leaderboards} lists open, non-English leaderboards and the models from our overview in Section~\ref{sec:statusquo} that they evaluate. 
We also report whether as of the current state, multilingual models or specialized target language models are in the lead in the size of up to 14B parameters.

\section{Generative Evaluation In Disguise: MMLU}\label{app:in_disguise}
Even though MMLU~\citep{hendrycks2020measuring} is by design a discriminative task (MCQA), it deserves to be discussed here as the most popular benchmark for multilingual models to-date because of the seemingly ease of evaluation (one of four options is correct). 
The original MMLU data has been translated automatically and with humans in various efforts with various translation tools (X-MMLU in Okapi work, \href{https://huggingface.co/datasets/openai/MMMLU}{MMMLU} (openAI), LLama3 report uses Google Translate for translation, GlobalMMLU~\citep{singh2024globalmmluunderstandingaddressing}), replicated in other languages~\citep{son2024kmmlumeasuringmassivemultitask, xuan2025mmluproxmultilingualbenchmarkadvanced}, analyzed and corrected (MMLU-Redux~\citep{gema2024mmlu}), sub-categorized (GlobalMMLU), extended (INCLUDE~\citep{romanou2024includeevaluatingmultilinguallanguage}), and critizices~\citep{balepur2025bestdescribesmultiplechoice}. However, since LLMs are generators by design, evaluation is not straightforward, therefore multiple approaches exist,\footnote{\url{https://huggingface.co/blog/open-llm-leaderboard-mmlu}} such as based on likelihood rankings of answers, or exact string matching. These details are rarely specified in multilingual MMLU evaluations, but may make the difference for system ranking~\citep{wei2024rethinking}.  % this paper only looks at English
MCQA tasks (the majority of them) can also be turned into a generative benchmark by stripping the answer options from the prompt and having a LLM judge decide whether the model's generation matches the correct answer option, possibly in comparison with another model's generation~\citep{wei2024rethinking}. The generative form of evaluation has so far not been explored multilingually.

\section{Multilingual vs Monolingual Models and Benchmarks}~\label{app:monolingual}
Individual language benchmarks and models often receive little recognition in multilingual LLM development. 
While extensive work on English monolingual LLMs is widely respected and adapted for other languages, monolingual or less massively multilingual models tend to be overlooked.
One challenge in evaluation arises when moving beyond monolingual to multilingual settings, as the language coverage of individual benchmarks and models often does not fully align. This mismatch can lead to benchmarks being deemed incomplete for certain models or, being overly extensive, unfairly penalizing models for languages they do not support.

One argument against specialization is the potential for \textit{sharing of information} in multilingual settings where knowledge learned from one language can benefit others \citep{conneau-etal-2020-emerging,artetxe-etal-2020-cross,desouza2021abilitymonolingualmodelslearn}, or general reasoning abilities transfer~\citep{chang2024goldfishmonolinguallanguagemodels}, especially with increased model sizes~\citep{chang2023multilingualitycurselanguagemodeling}.
Building monolingual models, however, provides the opportunity to specialize the model~\citep{chang2024goldfishmonolinguallanguagemodels}, including optimizing tokenization strategies \citep{chelombitko-komissarov-2024-specialized,zhao-aletras-2024-comparing} and tailoring the training data to the specific linguistic characteristics of the target language \citep{su2023welmwellreadpretrainedlanguage,abonizio2025sabia3technicalreport}.

\subsection{Experiment: Monolingual vs Multilingual Model Performance}
We compare a multilingual model, \textsc{Aya Expanse 8B}, with individual monolingual models on two open-ended generation tasks: general knowledge~\citep{singh-etal-2024-aya}, and a more challenging set of math, code, and reasoning questions~\citep{dang2024ayaexpansecombiningresearch}.
We cover a diverse set of monolingual models, including those pretrained from scratch exclusively on a single language, as well as models obtained by specialized finetuning of a multilingual model. Our selection also spans models with language-specific tokenizers versus general tokenizers, and models specialized for domains such as code or math in contrast to general-purpose language models. The languages include French\footnote{\url{https://huggingface.co/jpacifico/Chocolatine-3B-Instruct-DPO-v1.2}}, Hebrew\footnote{\url{https://huggingface.co/dicta-il/dictalm2.0-instruct}}, Chinese\footnote{\url{https://huggingface.co/01-ai/Yi-1.5-9B-Chat}}, Arabic\footnote{\url{https://huggingface.co/CohereLabs/c4ai-command-r7b-arabic-02-2025}}, and Japanese\footnote{\url{https://huggingface.co/llm-jp/llm-jp-3-7.2b-instruct3}}.
GPT4-o is used as a judge to evaluate the quality of generations in a pairwise comparison setting. 

Figure \ref{fig:mono_multi_winrates} shows that, in almost all cases, the multilingual model outperforms the monolingual counterparts in both evaluation sets. 
Previously on a smaller scale \citet{rust-etal-2021-good} compared mBERT \citep{devlin-etal-2019-bert} and pretrained monolingual BERT models across selected languages and showed that languages adequately represented in the multilingual model’s vocabulary exhibit little to no performance degradation compared to their monolingual counterparts. Now at a larger scale, ranging from 3B to 9B parameters, we observe an even stronger pattern of multilingual models outperforming their monolingual counterparts.

Aside from the challenges of properly configuring an entirely new model to generate coherent text in each new language, multilingual models also appear more powerful for open-ended generation tasks, consistently producing stronger outputs across languages.
However, this advantage may also stem from factors such as a higher number of experimental iterations, broader (rather than specialized) evaluation objectives, and the continuous updating and maintenance of multilingual models --- benefits that monolingual models, often developed in a more ``one-and-done'' style might lack.

\begin{figure}
    \centering
    \includegraphics[width=0.9\linewidth]{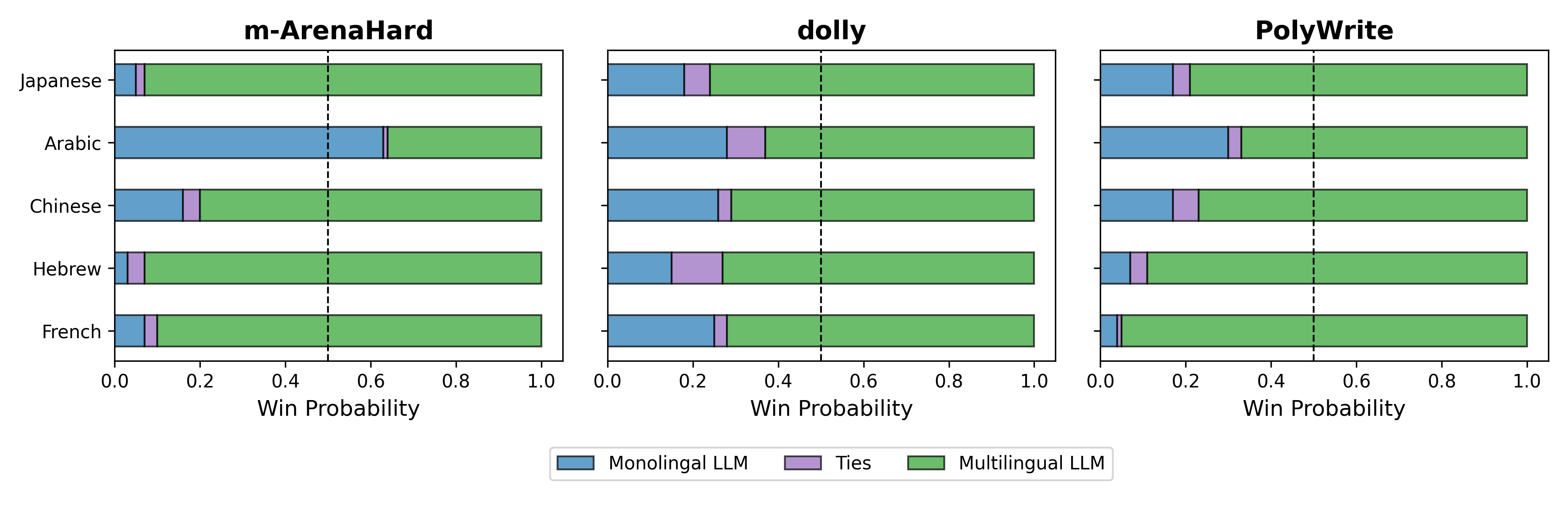}
    \caption{Comparing performance of a multilingual model (Aya Expanse 8B) with language-expert Monolingual models on general open-ended questions (dolly-translated-200, \citep{singh-etal-2024-aya} and creative writing prompts from PolyWrite \citep{ji2024emma500enhancingmassivelymultilingual}) and a more challenging set of math, code, and reasoning questions (m-ArenaHard, \citep{dang2024ayaexpansecombiningresearch})}
    \label{fig:mono_multi_winrates}
\end{figure}

\section{Win Rate Comparisons}\label{app:winrates}

\subsection{Sampling}\label{app:sampling}
For our win rate comparisons in ~\cref{sec:statistical_power}, we sample 5 generations from each model with ancestral sampling (temperature=1.0) as implemented in vLLM. We chose this setup because we did not want to tune temperatures individually for each model, nor was there any guide from either model provider how to set it in the best way.
In hindsight, we noticed that a lower temperature would have been beneficial for Qwen2.5, which explains why our Aya Expanse 8B winrates are more inflated than those in the Aya Expanse tech report~\citep{dang2024ayaexpansecombiningresearch}, especially for languages that had already lower quality. 
Upon communication with the authors, we found out that their evaluations were run with temperature=0.75, and we were able to confirm with spot checks of a few languages (from varying win rate buckets) that Qwen2 generations were of higher quality under that setup, see Table~\ref{tab:temp}. 
Win rates differences under different temperatures vary heavily, up to around 50 points in the most extreme case. For Japanese and Portuguese, even the directionality of the wins change: under temperatures 0.0 or 0.75, Qwen 2.5 wins overall, while under temperature=1.0, Aya Expanse wins overall.
Generally, this highlights how essential the documentation of decoding parameters is for replication.

\begin{table}[]
    \centering
    \begin{tabular}{llccc}
        \toprule
         \textbf{Language} & \textbf{Temperature} & \textbf{WR Aya Expanse} & \textbf{WR Qwen2.5} & WR $\Delta$\\
         \midrule
         hi 
         & 0.0 & 74.4 & 25.0 & 49.4 \\
         & 0.75 & 76.8 & 22.2 & 54.6 \\
         & 1.0 & 89.2 & 10.2 & \textbf{79.0}\\
         fa 
         & 0.0 & 59.0 & 40.6 & 18.4 \\
         & 0.75 & 71.0 & 28.2 & 42.8\\
         & 1.0 & 90.4 & 9.2 & \textbf{81.2}\\
         ja 
         & 0.0 & 44.8 & 54.8 & -10.0\\
         & 0.75  & 47.8 & 51.1 & -3.3*\\
         & 1.0 & 70.8 & 28.2 & \textbf{42.6}\\
         pt 
         & 0.0 & 43.6 & 55.6 & -12.0 \\
         & 0.75 & 41.8 & 57.0 & -15.2\\
         & 1.0 & 58.8 & 39.8 & \textbf{19.0}\\
         \bottomrule
    \end{tabular}
    \caption{The effect of temperature settings on win rates (WR, in \%) on mArenaHard for pairwise comparisons between Aya Expanse 8B and Qwen2.5 7B Instruct. Win rate differences are notably higher under temperature=1.0. Non-significant differences (95\% confidence interval) are marked with asterisk.}
    \label{tab:temp}
\end{table}

\subsection{LLM-as-a-Judge Prompting}\label{app:prompts}

\begin{table*}[]
    \centering
\begin{tabular}{p{1cm}|p{12cm}}    \toprule
        System & You are a helpful assistant whose goal is to select the preferred (least wrong) response for a given instruction in \texttt{language\_name}. \\
        Judge & Which of the following responses is the best one for the given instruction in \texttt{language\_name}? A good response should follow these rules:
    1) It should be in \texttt{language\_name},
    2) It should complete the request in the instruction,
    3) It should be factually correct and semantically comprehensible,
    4) It should be grammatically correct and fluent.\\
    & \\
  &  Instruction: \texttt{instruction}\\
  &  Response (A): \texttt{completion\_a}\\
  &  Response (B): \texttt{completion\_b}\\
 &   FIRST provide a concise comparison of the two responses. If one Response is better, explain which you prefer and why. If both responses are identical or equally good or bad, explain why.\\
  &  SECOND, on a new line, state exactly one of 'Response (A)' or 'Response (B)' or 'TIE' to indicate your choice of preferred response.\\
 &   Your response should use the format: Comparison: $<$concise comparison and explanation$>$ Preferred: $<$'Response (A)' or 'Response (B)' or 'TIE'$>$ \\
     \bottomrule
    \end{tabular}
    \caption{Prompts for LLM-as-a-judge evaluations}
    \label{tab:prompts}
\end{table*}

For LLM-as-a-judge evaluations, we use the prompts listed in ~\cref{tab:prompts} and randomize the order of model generations to prevent position bias. 
When using GPT4o as a judge, we use version 2024-11-20.

\subsection{Statistical Significance}
Preliminary experiments with GPT4o-mini (2024-07-18) for the setup described in ~\cref{sec:statistical_power} revealed that standard errors for win-rates were much higher than for GPT4o, so that even 500 examples are not enough for the differences to be significant in Chinese. 

%LLM_JUDGE_SYSTEM_PROMPT = "You are a helpful assistant whose goal is to select the preferred (least wrong) response for a given instruction in {language_name}.\n"
%LLM_JUDGE_PROMPT = (
%    "Which of the following responses is the best one for the given instruction in {language_name}? A good response should follow these rules:"
%    " 1) It should be in {language_name},"
%    " 2) It should complete the request in the instruction,"
%    " 3) It should be factually correct and semantically comprehensible,"
%    " 4) It should be grammatically correct and fluent.\n\n"
%    "Instruction: {instruction}\n"
%    "Response (A): {completion_a}\n"
%    "Response (B): {completion_b}\n"
%    "FIRST provide a concise comparison of the two responses. If one Response is better, explain which you prefer and why. If both responses are identical or equally good or bad, explain why.\n"
%    "SECOND, on a new line, state exactly one of 'Response (A)' or 'Response (B)' or 'TIE' to indicate your choice of preferred response.\n"
%    "Your response should use the format: Comparison: <concise comparison and explanation> Preferred: <'Response (A)' or 'Response (B)' or 'TIE'>"
%)}

\section{Instruction Wording}
\label{app:instruction_wording}

In this experiment, we use the German questions from Include 44 \citep{romanou2024include} test set containing localized multiple choice questions. MCQA test sets are usually evaluated with log-likelihood probability, however, when that is not possible, especially when comparing against models behind API, researchers reformulate the questions into instruction following.

We show how much the instruction can change final system ranking, thus opening a room for metric hacking. We design six different instructions in English and also translate them into German, all listed in \cref{fig_app:preprompt_wording}. The model outputs is then automatically parsed with regular expressions to select the proper answer. The final prompt contains the instruction followed with the question and list of all four answers and only the instruction is changed between experiments.

\Cref{tab_app:preprompt_wording} shows how different wording changes the final model accuracy for three different models.

\begin{table}[ht]
\centering
\begin{tabular}{lccc}
\hline
\textbf{} & \textbf{Aya Expanse 8B} & \textbf{Llama 3.1 8B} & \textbf{Gemma 2 9B} \\
\hline
EN 1 & 46.0\% & 50.4\% & 48.2\% \\
EN 2 & 44.6\% & 48.2\% & 47.5\% \\
EN 3 & 46.8\% & 48.2\% & 47.5\% \\
EN 4 & 48.9\% & 48.2\% & 49.6\% \\
EN 5 & 48.9\% & 50.4\% & 48.9\% \\
EN 6 & 42.4\% & 51.1\% & 51.1\% \\
DE 1 & 45.3\% & 51.8\% & 50.4\% \\
DE 2 & 48.2\% & 46.8\% & 51.1\% \\
DE 3 & 46.8\% & 48.9\% & 48.9\% \\
DE 4 & 46.8\% & 48.2\% & 48.2\% \\
DE 5 & 45.3\% & 51.8\% & 48.9\% \\
DE 6 & 43.9\% & 47.5\% & 48.2\% \\
\hline
\end{tabular}
\caption{Comparison of performance on German MCQA test set from Include 44 when using different instructions.}
\label{tab_app:preprompt_wording}

\end{table}

\begin{figure*}[ht]
\centering
\begin{Verbatim}[breaklines]
En 1: "Here's a multiple-choice question with answer options. Please respond with only the letter of the correct choice. Do not include any additional information in your answer.",
En 2: "Please examine the following multiple-choice question carefully and reply with just the letter corresponding to the correct answer. No additional text should be included in your response.",
En 3: "Select the correct option and reply with its corresponding letter only. Nothing else.",
En 4: "Here's a question for you! Just type the letter of the correct answer — do not provide any explanations or extra words.",
En 5: "Respond using only the letter of the correct answer. Do not add anything else.",
En 6: "Whatever you do, don't send anything besides the letter of the correct answer. No explanations, no extra words!",

De 1: "Hier ist eine Multiple-Choice-Frage mit Antwortoptionen. Bitte antworte nur mit dem Buchstaben der richtigen Auswahl. Füge deiner Antwort keine zusätzlichen Informationen hinzu.",
De 2: "Bitte überprüfe die folgende Multiple-Choice-Frage sorgfältig und antworte nur mit dem Buchstaben der richtigen Antwort. Füge deiner Antwort keinen zusätzlichen Text hinzu.",
De 3: "Wähle die richtige Option aus und antworte nur mit dem entsprechenden Buchstaben. Sonst nichts.",
De 4: "Hier ist eine Frage für dich! Gib nur den Buchstaben der richtigen Antwort ein – keine Erklärungen oder zusätzlichen Wörter.",
De 5: "Antworte nur mit dem Buchstaben der richtigen Antwort. Füge nichts Weiteres hinzu.",
De 6: "Was auch immer du tust, sende nichts außer dem Buchstaben der richtigen Antwort. Keine Erklärungen, keine überflüssigen Worte!"
\end{Verbatim}
\caption{English and German instructions for multiple choice question answering.}
\label{fig_app:preprompt_wording}

\end{figure*}

\section{Translation Effects}\label{app:translation}
\subsection{Experimental Setup}
The decoding setup is the same as for the win rate experiments described in Appendix~\ref{app:winrates}.
We choose the set of languages because they are in the common set of supported languages for our models of interest. 
We translate the original prompts via a pivot language back into the original language to simulate translation effects on the prompts. 
The pivot language is English for all languages except English, and Portuguese for English.
We translate with Google Translate, \textsc{NLLB-3.3}, \textsc{Aya Expanse 32B} and \textsc{Command A} to have a diverse mix of \mt{} and \mllm{} translators.
For \textsc{NLLB}, we split the prompt into individual sentences with the \texttt{sentence\_splitter} library,\footnote{\url{https://github.com/mediacloud/sentence-splitter}} before translation, and concatenate the translations.
We do not post-process the translations in any way, but we notice that the translations contain $<$unk$>$s, which can throw off generation models.

The translation template for \textsc{Aya Expanse 32B} and \textsc{Command A} is the following:
``You are a professional translator. Translate from \texttt{src\_language} into \texttt{target\_language}. Return nothing but the translation.'' We did not do extensive prompt tuning, but noticed that \textsc{Aya Expanse 32B} often answered prompts rather than translating them if we did not include an explicit instruction to only return the translation.

We evaluate outputs from \textsc{LLama3.1 8B} Instruct~\citep{grattafiori2024llama3herdmodels}, \textsc{Gemma2 9B}~\citep{gemmateam2024gemma2improvingopen}, \textsc{Aya Expanse 8B}, and \textsc{Qwen2.5 7B Instruct}~\citep{yang2024qwen2technicalreport} models.

\subsection{Translation Quality}
Table~\ref{tab:translation_quality} compares the corpus ChrF~\citep{popovic-2015-chrf} and \textsc{XCOMET-XL}~\citep{guerreiro-etal-2024-xcomet} scores of roundtrip translations, and reference-free \texttt{wmt23-cometkiwi-da-xl}~\citep{rei-etal-2023-scaling} scores for translation models on \texttt{aya-human-annotated} prompts.\footnote{Sacrebleu signature: \texttt{nrefs:1|case:mixed|eff:yes|nc:6|nw:0|space:no|version:2.5.1}.}
According to the roundtrip evaluation against the original prompt, Google Translate delivers the highest quality translations with a small margin over \textsc{Command A}, followed by \textsc{NLLB 3.3B} and \textsc{Aya Expanse 32B}.
%Average round-trip translation quality as measured by ChrF and XComet-XL across languages is highest for Google Translate (ChrF=69.17, XComet-XL=90.24) with a large margin to \textsc{Aya Expanse 32B} (ChrF=54.61, XComet-XL=58.79) and NLLB 3.3B (ChrF=59.03, XComet-XL=49.29). 
\textsc{NLLB} translates notably better into Arabic and English than \textsc{Aya Expanse 32B}, while  \textsc{Aya} translates better into Turkish and Chinese.

\begin{table}[]
    \centering
    \begin{tabular}{llccc}
    \toprule
        \textbf{Model} & \textbf{Language} & \textbf{ChrF } & \textbf{XComet} & \textbf{CometKiwi}\\
        \midrule
        Google Translate 
        & ar & 69.49  & 88.88 & 70.76\\
         & en & 86.52& 96.54 & 80.87\\
         & pt & 81.77 & 96.95 & 80.83\\
         & tr & 75.94 & 97.33 & 76.09\\
         & zh & 32.15& 88.52 & 70.68\\
         & \textit{Avg} & \textit{69.17} & \textit{93.65} & \textit{75.84}\\
         \midrule
         NLLB 3.3B
         & ar & 61.05 & 84.52 & 68.83\\
         & en & 79.88 & 95.52 & 81.12\\
         & pt & 75.29 & 95.71 & 78.77\\
         & tr & 60.66 & 92.60 & 71.01\\
         & zh & 18.29& 81.80 & 65.80\\
         & \textit{Avg} & \textit{59.03} & \textit{90.03} & \textit{73.11} \\
         \midrule
         Aya Expanse 32B 
         & ar & 35.11 &  80.83 & 66.01\\
         & en & 77.11 &  96.55 & 81.15\\
         & pt & 75.70 & 95.22 & 79.08\\
         & tr & 62.88 & 95.38 & 74.99\\
         & zh & 22.26& 84.89 & 68.64\\
         & \textit{Avg} & \textit{54.61} & \textit{90.57} & \textit{73.97}\\
         \midrule
         Command A
         & ar & 62.80 & 86.60 & 69.47\\
         & en & 82.72 & 96.32 & 78.25\\
         & pt & 67.06 & 96.00 & 79.63\\
         & tr & 81.90 & 97.12 & 75.88\\
         & zh & 37.12 & 87.99 & 67.09\\
         & \textit{Avg} & \textit{66.31} & \textit{92.80} & \textit{74.07}\\
         \bottomrule
    \end{tabular}
    \caption{Translation quality of prompt roundtrip translations of the Aya human annotated benchmark. ChrF and \textsc{XComet} are reference-based metrics computed for translations from pivot language to target language, and quality is estimated without references for the translation into the pivot language with \textsc{Comet-Kiwi}.}
    \label{tab:translation_quality}
\end{table}

%\begin{figure}
%    \centering
    %\includegraphics[width=0.5\linewidth]{figures/chrfs_prompt_translation.png}
    %\caption{Translation quality for prompt roundtrip translations on \texttt{aya-human-annotated} prompts.}
    %\label{fig:chrf_prompt_translation}
%\end{figure}

\subsection{Changes in Generation}
We want to measure how translation affects the generations.
For that purpose, we compute Spearman correlation between translation quality and the generation quality, relative to the untranslated version. Both quantities are computed with sentence-level ChrF. 
In Table~\ref{tab:similarities} we report these correlations for various \mllm{}s and translation models.
Overall, we find that correlations are always positive, meaning the better the prompt is translated, the closer the generation is to the generation for the untranslated prompt.
For translations from \textsc{NLLB} and \textsc{Aya} this correlation is stronger than for Google Translate, as they also have lower quality and thereby cause more changes to the prompts.
Across languages and translations, \textsc{Qwen} is the most susceptible to changes in the prompt. 

\begin{table}
 \resizebox{\textwidth}{!}{%
\begin{tabular}{lccccc}
\toprule
 \textbf{Model} & \textbf{NLLB 3.3B} & \textbf{Aya Expanse 32B} & \textbf{Command A} &\textbf{Google Translate} &  \textit{\textbf{Avg}}\\
\midrule
Qwen 2.5 7B Instruct & 0.31 & 0.58   & 0.23 &0.57 &\textbf{\textit{0.34}} \\
Gemma2 9B it & 0.41 & 0.43 & 0.34 & 0.28 & \textit{0.29}\\
Aya Expanse 8B & 0.37 & 0.35 & 0.28 & 0.26 & \textit{0.25} \\
Llama3.1 7B Instruct & 0.23 & 0.21 & 0.20 & 0.15 & \textit{0.16} \\

\textit{Avg }& \textbf{\textit{0.34}} & \textit{0.33} & \textit{0.23} & \textit{0.30} & \textit{0.24}\\
\bottomrule
\end{tabular}%
}
\caption{Pearson correlation between translation quality and generation quality, for several translation and generation models, averaged across languages. NLLB translations correlate the strongest with changes in generations, and \textsc{Qwen} seems most susceptible to translation artifacts in prompts.}
\label{tab:similarities}
\end{table}

\subsection{Changes in Win Rate}
\cref{tab:translated_winrate_app} lists the win rates for all languages in the comparison of \textsc{Aya Expanse 8B} vs \textsc{Gemma 2 9B} under GPT4o as a judge (2024-05-13).
We can see that the translation of prompts affects win rates across the bench, with a magnitude depending on the language and translation model.

\begin{table}[t]
    \centering
   % \resizebox{\textwidth}{!}{
    \begin{tabular}{lccccc}
        \toprule
         &  \cellcolor{gray!15} \textsc{og} & \multicolumn{4}{c}{Translator} \\
         & \cellcolor{gray!15}  & \textsc{nllb} & \textsc{gt} & \textsc{aya} & \textsc{Cmd A} \\
         \midrule
        ar & \cellcolor{gray!15}0.71 & 0.68 & 0.58 & \textbf{0.84} & 0.74 \\ 
        en & \cellcolor{gray!15}0.01* & -0.05* & 0.04* & \textbf{0.15} & 0.06* \\
        pt & \cellcolor{gray!15}0.08* & 0.09* & \textbf{0.13} &  0.02* & 0.00* \\
        tr & \cellcolor{gray!15} 0.29 & \textbf{0.43} & 0.42 & 0.40 & 0.42 \\
        zh & \cellcolor{gray!15}-0.21* & 0.16 & -0.19* & \textbf{0.20} & -0.08* \\
        %\cmidrule(lr){6-10} 
        \midrule
        \textit{Avg} & \cellcolor{gray!15}\textit{0.18} & \textit{0.26} & \textit{0.20} & \textbf{\textit{0.32}} & \textit{0.23} \\
       \bottomrule
    \end{tabular}%
  %  }
    \captionof{table}{Win-rate differences (\textsc{Aya Expanse 8B}-\textsc{Gemma2 9B}, *non-significant) on original prompts (\textsc{og}) vs translated prompts from various translation models. 
    Positive values mean that \textsc{Aya Expanse} wins, negative values mean that \textsc{Gemma} wins.
    }
    \label{tab:translated_winrate_app}
\end{table}

\section{Chatbot Arena Analysis}\label{app:chatbot_arena}

\subsection{Multilinguality}
The recently released 100k conversations and preferences collected between June and August 2024 (\href{https://huggingface.co/datasets/lmarena-ai/arena-human-preference-100k}{\texttt{lmarena-ai/arena-human-preference-100k}}) are more multilingual (54\%) than the previously released 33k data (\href{https://huggingface.co/datasets/lmsys/chatbot_arena_conversations}{\texttt{lmsys/chatbot\_arena\_conversations}}) from April to June 2023 that was 88\% English. 
The set of common languages with more than 200 prompts in each is: English, German, Spanish, French, Portuguese, Russian. 

\subsection{Ties}
Figure~\ref{fig:ties_chatbot_arena} shows the ratio of ties in human pairwise ratings for the five most prominent non-English languages from a total of 100k Chatbot Arena battles that were collected between June and August 2024 (\href{https://huggingface.co/datasets/lmarena-ai/arena-human-preference-100k}{\texttt{lmarena-ai/arena-human-preference-100k}}), and \cref{fig:ties_chatbot_arena_old} the same stats for the released battles from April to June 2023 (\href{https://huggingface.co/datasets/lmsys/chatbot_arena_conversations}{\texttt{lmsys/chatbot\_arena\_conversations}}).

\begin{figure}
    \centering
    \includegraphics[width=0.9\linewidth]{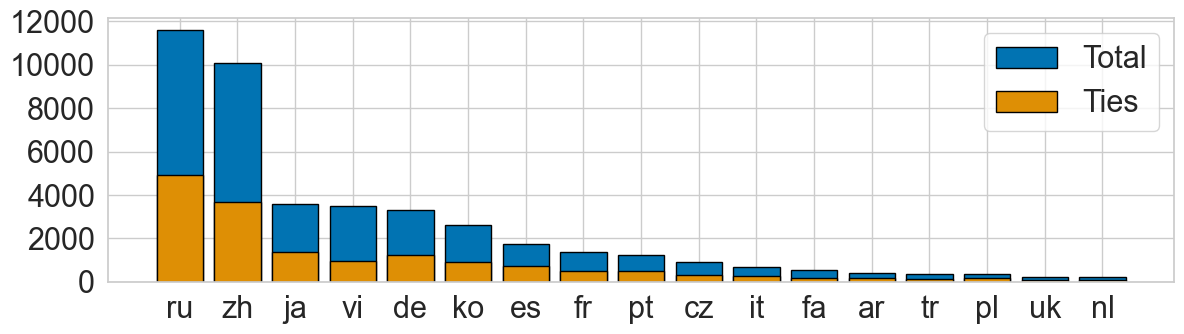}
    \caption{Number of total and tied Chatbot Arena battles (total 100k) for non-English languages with more than 200 prompts from 2024. 
    }
    \label{fig:ties_chatbot_arena}
\end{figure}

\begin{figure}
    \centering
    \includegraphics[width=0.5\linewidth]{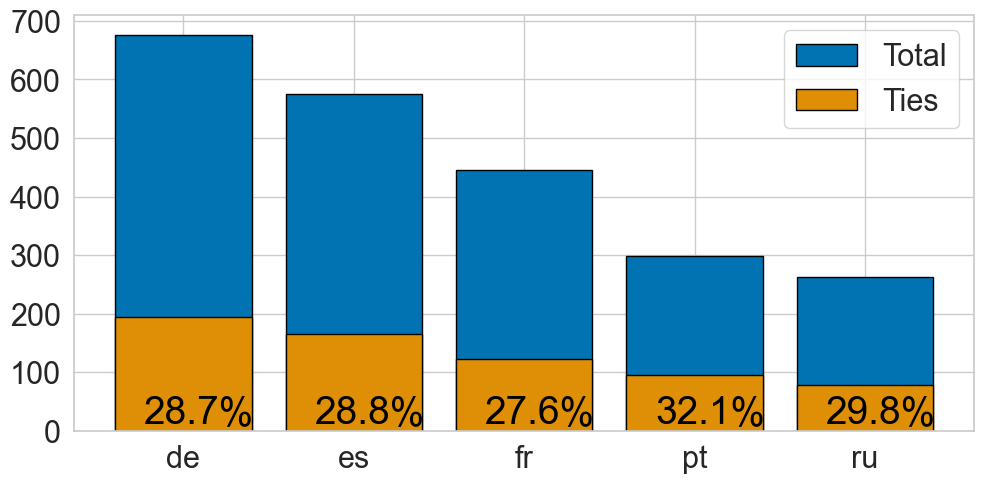}
    \caption{Number of total and tied Chatbot Arena battles (total 33k) for non-English languages with more than 200 prompts from 2023. }
    \label{fig:ties_chatbot_arena_old}
\end{figure}

\section{Checklist for Multilingual LLM Evaluation}\label{app:checklist}
\subsection{Evaluation Prompts}
\begin{enumerate}[label=$\square$]
    \item Are evaluation prompts representative samples of all languages included in the evaluation?
    \item Are evaluation prompts human-curated, localized or edited?
    \item If using model generated prompts: Have you analyzed the data for potential biases?
    \item If using translations: Have you estimated, reported, and attempted to optimize translation quality on this particular set of prompts?
\end{enumerate}

\subsection{Choice of Metrics}
\begin{enumerate}[label=$\square$]
    \item Are metrics adequate for all evaluated languages?
\end{enumerate}

\subsection{Statistical Testing}
\begin{enumerate}[label=$\square$]
    \item Does the evaluation include adequate statistical significance tests for all included languages?
    \item Does the evaluation include an estimate of statistical power?
    \item If using stochastic decoding, does it include estimates of sampling induced variance?
\end{enumerate}

\subsection{Aggregating Results Across Languages}
\begin{enumerate}[label=$\square$]
    \item Are the metrics comparable across languages?
    \item Is the aggregation of results disproportionately influenced by any outliers?
    \item Are language support differences taken into consideration when aggregating results across languages?
    \item Are language support differences documented with the results?
    \item Are task- and language-specific scores reported?
\end{enumerate}

\subsection{Qualitative Insights}
\begin{enumerate}[label=$\square$]
    \item Are quantitative metrics accompanied by qualitative error analyses?
    \item Are differences in metrics due to meaningful distinctions rather than incidental artifacts?
\end{enumerate}

\subsection{Reproducibility}
\begin{enumerate}[label=$\square$]
    \item Are the results calculated with standardized pipelines?
    \item Is the evaluation code released?
    \item Are exact evaluation prompts and format published?
    \item Are model outputs released?
    \item Are prompt-level evaluation scores released?
    \item Are metric hyperparameters documented?
    \item Are model versions documented?
\end{enumerate}

\subsection{Enabling Meta-Evaluation}
\begin{enumerate}[label=$\square$]
    \item Are any human evaluations (consensually) released?
\end{enumerate}

\end{document}